\documentclass{article} 
\usepackage{iclr2027_conference,times}

\usepackage{amsmath,amsfonts,bm}

\def\eqref#1{equation~\ref{#1}}

\def\1{\bm{1}}

\DeclareMathAlphabet{\mathsfit}{\encodingdefault}{\sfdefault}{m}{sl}
\SetMathAlphabet{\mathsfit}{bold}{\encodingdefault}{\sfdefault}{bx}{n}

\newcommand{\R}{\mathbb{R}}

\usepackage{amsmath}
\usepackage{amssymb}
\usepackage{algorithm}
\usepackage{algorithmic}
\usepackage{graphicx}
\usepackage{booktabs}
\usepackage{comment}

\usepackage{hyperref}
\usepackage{url}
\usepackage{amsmath,amssymb,amsfonts}
\usepackage{algorithm}
\usepackage{algorithmic}
\usepackage{multirow}
\usepackage{array}
\usepackage{xcolor}
\usepackage[table]{xcolor}
\usepackage{verbatim}
\usepackage{fvextra}
\usepackage{url}
\usepackage{hyperref}
\usepackage{natbib}
\usepackage{type1cm}

\newcommand{\tablefontsize}{\fontsize{7.5pt}{8.75pt}\selectfont}

\title{AVC-Reward: Event-Centric Audio-Video \\ Consistency Evaluation}

\author{Yuruo Wu$^{1}$\thanks{Equal contribution. Email: yuruowu@zju.edu.cn.},\; 
Kaihang Pan$^{1}$\footnotemark[1],\; Kai Shen$^{1}$\footnotemark[1],\; Jiacheng Li$^{2}$,\; Shuyuan Tu$^{3}$,\; \\
\textbf{Li Tuoyu$^{1}$,\; Juncheng Li$^{1}$\thanks{Corresponding author. Email: junchengli@zju.edu.cn.},\; Siliang Tang$^{1}$} \\
\vspace{-0.2em}\\
$^{1}$Zhejiang University, $^{2}$HiThink Research, 
$^{3}$Fudan University}

\iclrpreprintcopy
\begin{document}
\maketitle
\begin{abstract}
As joint audio-video generation models become more capable, they can generate audio-visual scenes exhibiting three forms of complexity: \textbf{multi-event composition}, \textbf{multi-layer audio mixture}, and \textbf{asymmetric cross-modal relations}.
However, current evaluations largely reuse metrics from earlier simple and unidirectional tasks. They typically evaluate an audio-video clip as a whole and rely on simple predefined relations, making them unsuitable for these increasingly complex scenes.
We propose an \textbf{event-centric} evaluation framework that disentangles audio and visual events at the event level, analyzes each event independently, cross-validates relations across modalities, and aggregates the judgments into a \textbf{coarse-to-fine, three-level consistency metric}.
The metric assesses global narrative harmony, event-level temporal synchronization, and fine-grained attribute alignment, providing a general formulation for audio-video consistency evaluation across diverse generation settings and cross-modal relations.
We instantiate the metric as \textbf{AVC-Reward}, a trainable audio-video reward model.
To evaluate the agreement between automatic judgments and human perception across generation paradigms and relation complexities, we construct AVC-Benchmark, which is primarily built from diverse positive and negative samples generated by various types of audio-video generation models.
AVC-Reward achieves state-of-the-art alignment with human judgment among automatic methods, provides interpretable rationales that localize failures, and improves the audio-video consistency of a joint generation model when used as an online reinforcement-learning signal.
Project page: \url{https://tsuki0512.github.io/AVC-Reward.github.io/}
\end{abstract}

\section{Introduction}
\label{sec:intro}

Audio-video generation has evolved from unidirectional driving toward joint generation with increasingly strong generative capabilities. Earlier audio-driven video generation (e.g., Wav2Lip~\citep{56wav2lip}) and video-to-audio generation (e.g., HunyuanVideo-Foley~\citep{25hunyuanfoley}) were centered on more constrained consistency relations: the former mainly focused on lip synchronization~\citep{01syncnet}, while the latter mainly focused on simple temporal alignment~\citep{07desync}. In contrast, joint generation removes the constraint of a fixed modality, with recent models such as Veo~3~ \citep{03veo3}, MiniMax H3~\citep{57minimaxh3} and Seedance~2.0~\citep{45seedance} extending this paradigm to richer and more complex scenes.

These richer audio-visual scenes exhibit three forms of complexity: \textbf{multi-event composition}, \textbf{multi-layer audio mixture}, and \textbf{asymmetric cross-modal relations}. First, a clip can contain multiple events that together form a complete scene, with several events potentially occurring at the same time. Second, its audio track can mix physical scene sounds with post-produced audio such as voice-over and background music, which instead support the video's narrative or atmosphere. Third, audio and visual events need not follow a simple one-to-one relation: valid audio events such as off-screen sounds may lack direct visible visual evidence.

However, \textbf{evaluation methodologies have not kept pace} with this evolution: the richer audio-visual content produced by joint models has made consistency evaluation more demanding.
Current evaluations largely reuse metrics from earlier simple and unidirectional tasks, where one modality was fixed and only a limited relation such as lip--speech or sound--action timing needed to be checked; accordingly, they often evaluate an audio-video clip as a whole and rely on simple predefined relations such as semantic similarity or temporal synchronization.
\textbf{Multi-event composition} makes such clip-level evaluation unreliable because different event relations are entangled; for example, ImageBind-based metrics~\citep{14imagebind,34avhscore,04javisdit} encode the audio and video within a clip or temporal window into a shared embedding space and measure their overall similarity, which can remain high when dominant semantics match despite a local event inconsistency.
\textbf{Multi-layer audio mixture} further complicates evaluation: metrics such as Synchformer~\citep{07desync} predict an audio--video temporal offset from the mixed soundtrack without separating its layers. When physical sounds are mixed with voice-over or background music, sounds that should align with visible actions are evaluated together with audio that does not require such temporal correspondence, which can lead to an inaccurate consistency score.
\textbf{Complex cross-modal relations} further challenge predefined matching rules; for example, AV-Align~\citep{15avalign} matches audio onsets with visual-motion peaks and may therefore penalize a valid off-screen sound that lacks direct visual evidence.

Therefore, we need a more general audio-video consistency evaluator that can handle diverse event relations across different generation paradigms, particularly in modern joint generation.
Moreover, we believe that such an evaluation model can serve as a training signal to directly improve generation models, closing the loop of \textbf{``generate--evaluate--optimize.''}

\begin{figure}[t]
    \centering
    \includegraphics[width=0.9\linewidth]{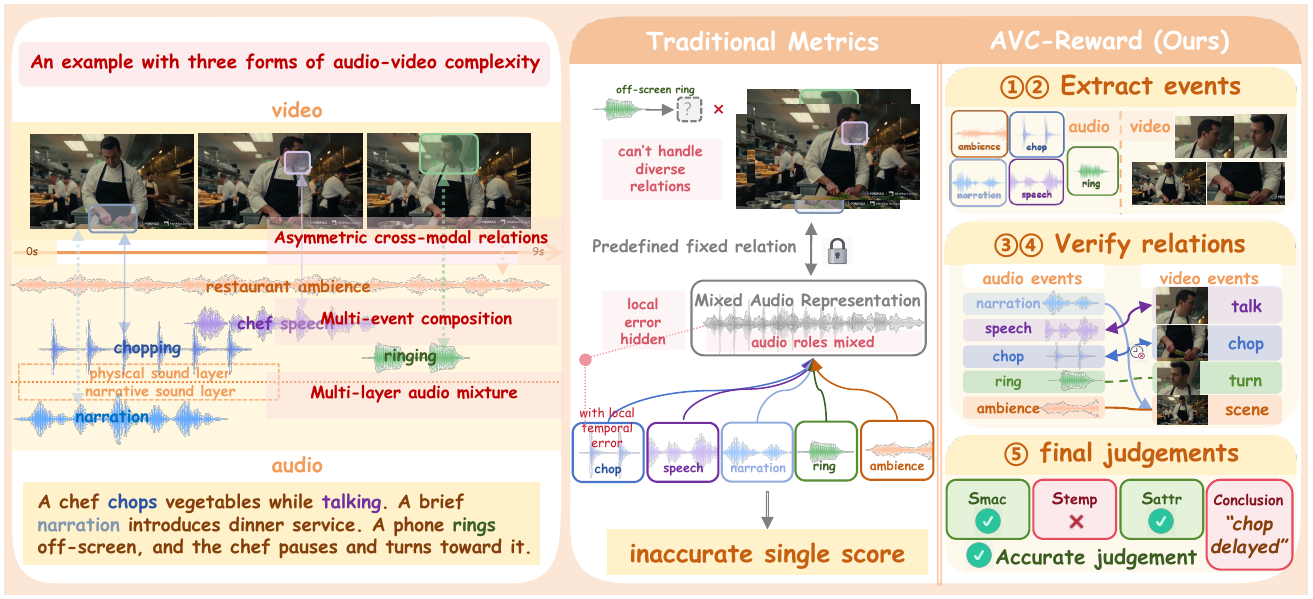}
    \caption{Overview of the proposed event-centric audio-video consistency evaluation framework.}
    \label{fig:main_overview}
\end{figure}

To meet this need, we propose an \textbf{event-centric} evaluation perspective, in which the audio and video of each generated clip are first disentangled into individual events and analyzed independently before the final judgments.
Under this perspective, we first separately extract and analyze audio and visual events, then cross-validate their relations across modalities, and finally aggregate the resulting judgments from coarse to fine.
We formulate this final consistency assessment as \textbf{a coarse-to-fine, three-level consistency metric}.
\textbf{Global narrative harmony} assesses whether the audio and video form a coherent scene and narrative.
\textbf{Event-level temporal synchronization} establishes event relations, checks event presence and temporal alignment, and marks audio events without visual evidence.
\textbf{Fine-grained attribute alignment} verifies the attributes of matched events and assesses whether marked audio is plausibly an off-screen sound, voice-over, or background music.
By disentangling individual events, distinguishing different audio sources, and explicitly validating their cross-modal relations, this metric can separately assess overlapping events, handle audio layers with different roles, and accommodate valid asymmetric relations, as illustrated in Fig.~\ref{fig:main_overview}.

Based on this event-level perspective, we implement the above three-level metric as a trainable audio-video reward model, named \textbf{AVC-Reward}, initialized from Qwen3-Omni-Thinking~\citep{09qwen3omni}. AVC-Reward implements the metric through a five-stage structured reasoning chain consisting of audio(/video)-only analysis, cross-modal verification, and overall judgment.
Its training data includes audio-video samples from different generation paradigms and models, with annotations from multiple expert MLLMs and human annotators, while its two-stage training procedure uses supervised fine-tuning (SFT) to cold-start the structured evaluation process and group relative policy optimization (GRPO)~\citep{10grpo} to align the scoring with human preferences.
Meanwhile, to measure the agreement between audio-video consistency metrics and human perception across generation paradigms and relation complexities, we construct an audio-video consistency benchmark, \textbf{AVC-Benchmark}. To our knowledge, it is the first audio-video consistency benchmark primarily  built from diverse samples generated by various types of audio-video generation models.

We evaluate the reward model from two complementary perspectives. For \textbf{evaluation effectiveness}, it achieves \textbf{state-of-the-art alignment with human judgment} on AVC-Benchmark, and unlike a black-box scalar, it provides \textbf{interpretable, structured natural-language rationales} that localize where a generation fails (e.g., a temporally misaligned event or a hallucinated sound). For \textbf{usability as a reward}, applying it as a training signal for the online reinforcement-learning fine-tuning~\citep{51omninft} of a joint generation model LTX-2~\citep{49ltx2} measurably improves the audio-video consistency of generated outputs, demonstrating that it closes the optimization loop.
Our contributions are summarized as follows:
\begin{itemize}
    \item We formalize \textbf{audio-video consistency} across generation paradigms with an \textbf{event-centric, three-level metric} designed to handle complex generated scenes.
    \item We instantiate this metric as a trainable \textbf{audio-video reward model} built on the Qwen3-Omni-Thinking~\citep{09qwen3omni} model with rigorous event-level verification.
    \item We construct \textbf{AVC-Benchmark} for reproducible human-alignment evaluation, and validate the reward model both as an interpretable automatic evaluator and as an effective training signal that \textbf{improves the audio-video consistency of a joint generation model}.
\end{itemize}

\section{Related Work}
\label{sec:related}

\paragraph{Joint Audio-Video Generation.}
\label{sec:related_gen}
Audio-video generation first developed through audio-driven video generation~\citep{56wav2lip,36echomimic,24hallo3,08omnihuman} and video-to-audio generation~\citep{02foleycrafter,39mmaudio,25hunyuanfoley}, where one modality is fixed.
Joint audio-video generation (JAVG) removes this constraint: early works either learned a joint generative process~\citep{40mmdiffusion} or coordinated pretrained unimodal generators~\citep{41seeingandhearing}.
Native JAVG then advanced toward open-ended generation, finer synchronization, and richer audio content, as represented by JavisDiT~\citep{04javisdit}, Ovi~\citep{13ovi}, UniVerse-1~\citep{42universe1}, and LTX-2~\citep{49ltx2}.
JAVG has further reached industrial-scale deployment through systems such as Veo~3~\citep{03veo3}, Seedance~2.0~\citep{45seedance}, and MiniMax H3~\citep{57minimaxh3}, enabling the generation of more complex audio-visual scenes with multi-event composition, {multi-layer audio mixtures}, and complex cross-modal relations, which makes audio-video consistency evaluation more demanding.

\begin{table}[t]
\centering
\setlength{\tabcolsep}{2pt}
\caption{Comparison of audio-video consistency metrics for challenges exposed by modern JAVG.}
\label{tab:related_metrics}
\tablefontsize

\begin{tabular*}{\linewidth}{@{}l@{\extracolsep{\fill}}lll@{}}
\toprule
\textbf{Method} &
\textbf{Scoring Unit} &
\textbf{Audio layers} &
\textbf{Relation Evaluated} \\
\hline

SyncNet~\citep{01syncnet}
& Mouth--speech window
& Mixed
& Lip--speech synchrony \\

AV-Align~\citep{15avalign}
& Audio / motion peaks
& Mixed
& Motion--sound timing \\

DeSync~\citep{07desync}
& AV temporal window
& Mixed
& Temporal offset \\

ImageBind~\citep{14imagebind}
& Clip
& Mixed
& Global semantic similarity \\

AVHScore~\citep{34avhscore}
& Frame $\times$ whole audio
& Mixed
& Frame--audio similarity \\

JavisScore~\citep{04javisdit}
& Temporal window
& Mixed
& Window-level AV similarity \\

PEAVS~\citep{16peavs}
& Clip
& Mixed
& Predefined AV failures \\

AVBench~\citep{yang2026avbenchhumanalignedautomatedevaluation}
& Clip
& Mixed
& Predefined AV failures \\

AGAV-Rater~\citep{pmlr-v267-cao25f}
& Clip
& Mixed
& Generative AV consistency \\

Beyond Time Shifts~\citep{qian2026beyond}
& Clip
& Mixed
& Generative AV consistency \\  \hline

\textbf{Ours}
& \textbf{Event}
& \textbf{Decoupled}
& \textbf{Event-dependent relation} \\

\bottomrule
\end{tabular*}
\end{table}

\paragraph{Audio-Video Consistency Metrics.}
\label{sec:related_metric}
Existing audio-video consistency metrics and their limitations under modern joint audio-video generation are summarized in Table~\ref{tab:related_metrics}. 
Early metrics target specific consistency relations: SyncNet measures lip--speech synchrony, AV-Align matches motion and audio peaks, and DeSync predicts temporal offsets.
ImageBind, AVHScore, and JavisScore measure cross-modal similarity at clip, frame--audio, or temporal-window granularity.
PEAVS and AVBench learn human preferences from manually created errors.
AGAV-Rater and Beyond Time Shifts further learn from real V2A outputs.
Recent analysis further shows that existing metrics focus on different aspects of consistency and exhibit low cross-metric agreement~\citep{sharma2026what}. Together, these methods remain tied to specific tasks or relations, making them insufficient for general audio-video consistency evaluation under the complexity of modern joint generation.
\section{Method}
\label{sec:method}

\subsection{Evaluation Overview}
\label{sec:overview}

The complexities exposed by modern joint generation introduce three requirements for evaluation.
Multi-event composition calls for \textbf{event disentanglement}, so that individual and overlapping events can be examined separately.
Multi-layer audio mixture requires the evaluator to \textbf{distinguish the roles of different audio events}, as different audio sources can have different relations to the video.
Asymmetric and diverse cross-modal relations further require \textbf{establishing and verifying event-dependent cross-modal relations} for individual events.

Meeting these requirements requires an evaluator capable of jointly understanding audio-visual content and reasoning over different events and their relations. MLLMs with audio-video understanding capabilities can jointly perform these tasks within a unified reasoning process, making them a suitable foundation for audio-video consistency evaluation.

Moreover, for more general and accurate evaluation, we argue that the judgments of the MLLM should be organized into an \textbf{event-centric} metric with \textbf{three coarse-to-fine levels}.
\textbf{Global narrative harmony} first checks whether the audio and video are coherent across the full clip.
\textbf{Event-level temporal synchronization} and \textbf{fine-grained attribute alignment} then move to the event level.
They respectively assess the temporal alignment of each event and the consistency of audio-visual attributes within that event.
Through event-level evaluation and detailed, independent verification of each event's cross-modal relations, the metric addresses the higher evaluation demands above.

With this evaluation design in place, Section~\ref{sec:protocol} defines the event-centric protocol and three-level metric, Section~\ref{sec:data} constructs the corresponding training supervision, and Section~\ref{sec:train} learns AVC-Reward through SFT and GRPO.

\subsection{Event-Centric Evaluation Protocol}
\label{sec:protocol}

Following Section~\ref{sec:overview}, we define three-level consistency scores.
\textbf{Global narrative harmony} ($S_{\mathrm{mac}}$) measures agreement in narrative content, entities, environment, and affective tone.
\textbf{Event-level temporal synchronization} ($S_{\mathrm{temp}}$) measures event presence and the alignment of onset, offset, and duration under the appropriate cross-modal relations.
\textbf{Fine-grained attribute alignment} ($S_{\mathrm{attr}}$) measures attribute compatibility within verified relations, including sound--material agreement, lip-shape agreement with speech, and plausible responses to off-screen sounds.
All three scores lie in $[0,1]$, with higher values indicating greater consistency.

To compute these scores, we guide the MLLM through a five-stage reasoning protocol that specifies the evidence and judgments required at each stage.
Given an observed audio-video pair $\mathcal{P}=(\mathcal{A},\mathcal{V})$, the first two stages direct the model to analyze each modality independently:
\begin{equation}
    \mathcal{H}^{A}=(\mathcal{G}^{A},\mathcal{E}^{A}), \qquad
    \mathcal{H}^{V}=(\mathcal{G}^{V},\mathcal{E}^{V}),
\end{equation}
where $\mathcal{H}$ denotes an analysis record, $\mathcal{G}$ denotes global scene and content information, and $\mathcal{E}$ denotes the event set.
Each event records its temporal span and attributes; audio events additionally record their source role.
The next two stages verify the extracted events against the other modality:
\begin{equation}
    \mathcal{R}^{A\rightarrow V}
    =F_{A\rightarrow V}(\mathcal{H}^{A},\mathcal{V}), \qquad
    \mathcal{R}^{V\rightarrow A}
    =F_{V\rightarrow A}(\mathcal{H}^{V},\mathcal{A}).
\end{equation}
Here, $F_{A\rightarrow V}$ checks visual evidence for audio events, and $F_{V\rightarrow A}$ checks expected sounds for visual events.
The required evidence depends on each audio event's role.
The output sets $\mathcal{R}^{A\rightarrow V}$ and $\mathcal{R}^{V\rightarrow A}$ contain verification records $r_k=(e_i^{A},e_j^{V},\rho_k)$.
Here, $e_i^{A}$ and $e_j^{V}$ denote related audio and visual events, and $\rho_k$ specifies their relation and verification evidence.
Either event may be absent when evidence is missing or the relation is asymmetric.

Let $\mathcal{R}=\mathcal{R}^{A\rightarrow V}\cup\mathcal{R}^{V\rightarrow A}$.
The final stage scores global consistency from the two global descriptions and event consistency from the verified evidence:
\begin{equation}
    S_{\mathrm{mac}}=\Phi_{\mathrm{mac}}(\mathcal{G}^{A},\mathcal{G}^{V}),\qquad
    S_{\mathrm{temp}}=\Phi_{\mathrm{temp}}(\mathcal{R}),\qquad
    S_{\mathrm{attr}}=\Phi_{\mathrm{attr}}(\mathcal{R}).
\end{equation}
The $\Phi$ functions represent the MLLM's judgments under the three scoring criteria. Scoring proceeds from global semantics to event presence and timing, followed by attribute checks on relations that pass temporal verification. The three scores remain separate.
When an overall score is required, we use their arithmetic mean as $S_{\mathrm{AVC}} =\frac{1}{3}(S_{\mathrm{mac}}+S_{\mathrm{temp}}+S_{\mathrm{attr}})$.

We use $S_{\mathrm{AVC}}$ as the default overall score in our experiments.
Full event definitions and scoring criteria are given in App.~\ref{app:metric}.

\begin{figure}[t]
    \centering
    \includegraphics[width=\linewidth]{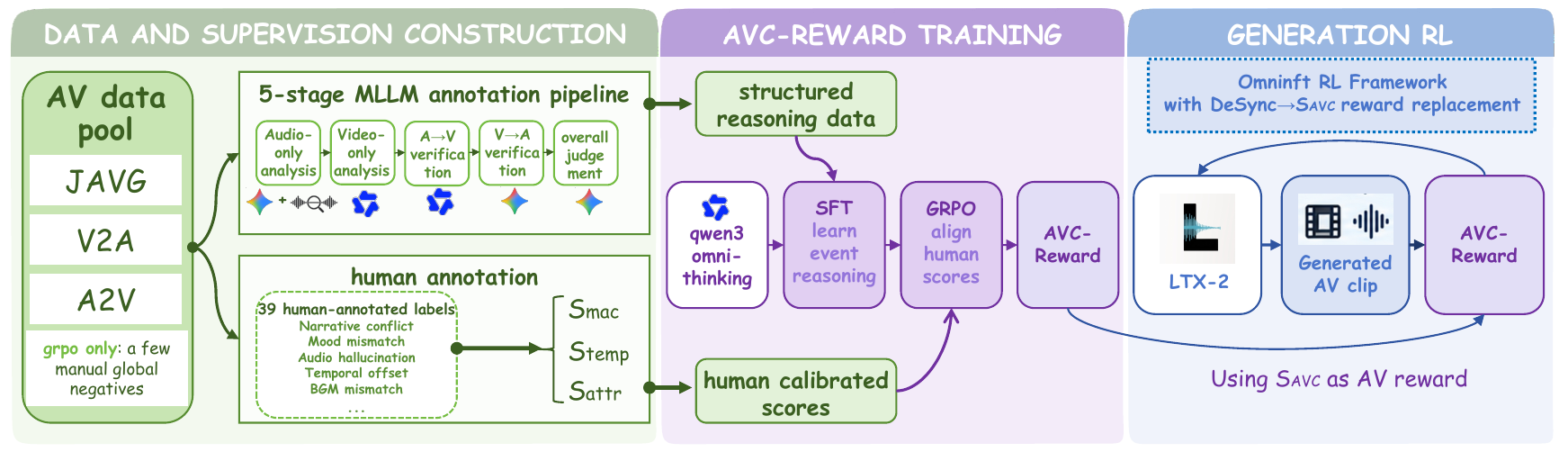}
    \caption{Overview of data construction, reward-model training, and generation-side optimization.}
    \label{fig:data_training_overview}
\end{figure}

\subsection{Training Data Construction and Annotation}
\label{sec:data}

Our training data includes outputs from existing joint audio-video, audio-driven video, and video-to-audio generation models. We further add a small number of manually edited global-narrative negatives in GRPO to balance the distribution of failure cases. The overall pipeline is illustrated in Fig.~\ref{fig:data_training_overview}.
We provide two complementary forms of supervision: structured reasoning data teaches the model to follow the event-centric evaluation protocol for SFT, while human-calibrated data aligns its scores with human judgment for GRPO.
The complete data sources and calibration details are provided in App.~\ref{app:taxonomy}.

\paragraph{Structured reasoning data.}
Following the five-stage protocol in Section~\ref{sec:protocol}, the annotation organizes modality-specific event analysis, bidirectional cross-modal verification, and the final coarse-to-fine judgment into a structured reasoning process. We construct the annotations using Gemini 2.5 Pro~\citep{19gemini25} and Qwen3-VL-Max~\citep{21qwen3vl} as complementary annotators, with a silence detector providing auxiliary temporal cues. Each stage produces one structured record.
The five structured records are concatenated and rewritten by Qwen3-235B~\citep{22qwen3} into a single SFT target,
$\langle\texttt{think}\rangle\,\mathcal{C}\,
\langle\texttt{/think}\rangle
\{S_{\mathrm{mac}},S_{\mathrm{temp}},S_{\mathrm{attr}},\kappa\}$,
where $\mathcal{C}$ preserves the five-stage reasoning order. After filtering, approximately 8k samples are retained for SFT.
These samples provide explicit supervision for learning event-specific evidence and event-dependent cross-modal relations under the three-level metric.
Complete reward-model prompts and input--output examples are provided in App.~\ref{app:case}.

\paragraph{Human annotation and score calibration.}
The structured annotation pipeline provides scalable reasoning supervision, but its scores are limited by the perceptual capability of the annotation models and can accumulate errors across stages. We therefore introduce human annotations to calibrate the scores toward human judgment.

We design a 39-label taxonomy over the three consistency dimensions, covering normal states and common failure types. Annotators assign applicable labels to each sample, grounding their judgments in explicit consistency evidence.
We further construct a multi-annotator subset with both label annotations and independently assigned dimension scores. Based on this subset, Qwen3-235B~\citep{22qwen3} maps each human-verified label record to $(S_{\mathrm{mac}},S_{\mathrm{temp}},S_{\mathrm{attr}})$ and a diagnostic conclusion $\kappa$. This calibration produces approximately 10k human-calibrated samples for GRPO.

\subsection{AVC-Reward Training}
\label{sec:train}

We initialize AVC-Reward from Qwen3-Omni-Thinking~\citep{09qwen3omni} and train it in two stages.

\paragraph{SFT cold start.}
We fully fine-tune the model on the approximately 8k structured reasoning samples from Section~\ref{sec:data}, using the complete five-stage reasoning and assessment as the training target.

\paragraph{GRPO for human alignment.}
After SFT cold start, we apply GRPO~\citep{10grpo} on a 10k rollout set constructed from the human-calibrated data. The reward jointly preserves the required evaluation structure and aligns the three consistency scores with human judgment. Specifically, a structure reward $r_{\mathrm{struct}}$ enforces the five-stage reasoning and assessment format. A score distillation reward
\(
r_{\mathrm{distill}}=\frac{1}{3}\sum_{i=0}^{2}\exp[-w_i(\mathrm{pred}_i-\mathrm{ref}_i)^2]
\)
measures agreement with the human-calibrated scores, and the total reward is
\(r_{\mathrm{total}}=(2r_{\mathrm{struct}}+r_{\mathrm{distill}})/3\),
where $i=0,1,2$ corresponds to $(S_{\mathrm{mac}},S_{\mathrm{temp}},S_{\mathrm{attr}})$.
\section{Benchmark}
\label{sec:benchmark}

\paragraph{Motivation.}
We construct \textbf{AVC-Benchmark} to evaluate how well automatic audio-video consistency metrics align with human perception in realistic generation scenarios.

\paragraph{Benchmark composition and evaluation target.}
AVC-Benchmark is the first benchmark, to our knowledge, that primarily uses various types of model-generated audio-video data to evaluate the alignment between automatic consistency metrics and human perception.
It contains samples from A2V, V2A, and JAVG models, together with a small set of controlled negatives added to balance the failure-case distribution and broaden the coverage of audio-video inconsistencies.
Each sample is independently annotated by three annotators, providing human ground-truth scores on the three consistency dimensions defined in Section~\ref{sec:protocol}.
We evaluate each metric against these human scores using pairwise accuracy under different score-difference thresholds, Spearman's rank correlation coefficient (SRCC), and Pearson's linear correlation coefficient (PLCC).

\paragraph{Statistics.}
Detailed benchmark statistics are provided in App.~\ref{app:benchmark_statistics}.
In general, most samples retain high global scene and narrative consistency, providing a meaningful basis for evaluating finer temporal and attribute-level relations; accordingly, the latter two dimensions span much broader score ranges.
The annotated failures further cover global narrative, affective, and environmental conflicts; temporal errors across physical events, speech, and music; and fine-grained mismatches in material, morphology, and causal response.

\section{Experiments}
\label{sec:experiments}

We evaluate AVC-Reward as both an automatic evaluator and a training reward for joint generation.
Section~\ref{sec:exp_reward} evaluates its agreement with human judgments; Section~\ref{sec:exp_gen} studies generation-side optimization; Section~\ref{sec:exp_human} reports human evaluation; Section~\ref{sec:exp_case} presents case studies; and Section~\ref{sec:exp_ablation} examines the main design choices.
\subsection{Reward Model Evaluation}
\label{sec:exp_reward}
\begin{table}[t]
\centering
\caption{Main results on AVC-Benchmark (706 samples). $\Delta$-Pairwise ($\Delta\!>\!0.2$) and PLCC are the primary metrics. \textbf{Best} automatic result in bold; \emph{Human Upper Bound} is the inter-annotator ceiling.}
\label{tab:bench}
\tablefontsize
\setlength{\tabcolsep}{2.5pt}
\renewcommand{\arraystretch}{0.96}
\begin{tabular}{lcccc}
\toprule
Method & Pairwise $\uparrow$ & $\Delta$-Pairwise $\uparrow$ & PLCC $\uparrow$ & SRCC $\uparrow$ \\
\hline
\multicolumn{5}{l}{\textit{Timing-based synchronization}} \\
AV-Align~\citep{15avalign}            & 0.577 & 0.674 & 0.301 & 0.215 \\
DeSync~\citep{07desync}               & 0.521 & 0.580 & 0.280 & 0.061 \\
\hline
\multicolumn{5}{l}{\textit{Embedding similarity}} \\
ImageBind~\citep{14imagebind}         & 0.556 & 0.625 & 0.219 & 0.154 \\
JavisScore~\citep{04javisdit}         & 0.554 & 0.611 & 0.180 & 0.147 \\
AVHScore~\citep{34avhscore}           & 0.540 & 0.595 & 0.171 & 0.110 \\
\hline
\multicolumn{5}{l}{\textit{Learned cross-modal consistency evaluators}} \\
PEAVS~\citep{16peavs}                 & 0.491 & 0.505 & 0.075 & -0.026 \\
AVSyncEvaluator~\citep{qian2026beyond}               & 0.502 & 0.554 & 0.108 & 0.007 \\
AV-Consistency Evaluator~\citep{yang2026avbenchhumanalignedautomatedevaluation}           & 0.562 & 0.664 & 0.286 & 0.173 \\
AGAV-Rater~\citep{pmlr-v267-cao25f}                       & 0.580 & 0.646 & 0.261 & 0.226 \\
\hline
\multicolumn{5}{l}{\textit{Zero-shot general MLLM judges}} \\
Gemini-3.1-Pro~\citep{20gemini3}      & 0.542 & 0.616 & 0.246 & 0.118 \\
Qwen3-Omni-Instruct~\citep{09qwen3omni} & 0.548 & 0.589 & 0.257 & 0.136 \\
Qwen3-Omni-Thinking~\citep{09qwen3omni} (base) & 0.555 & 0.605 & 0.347 & 0.154 \\
\hline
\textbf{AVC-Reward (Ours)}       & \textbf{0.586} & \textbf{0.728} & \textbf{0.494} & \textbf{0.244} \\
\hline
\textit{Human Upper Bound}            & \textit{0.756} & \textit{0.920} & \textit{0.781} & \textit{0.497} \\
\bottomrule
\vspace{-1.5em}
\end{tabular}
\end{table}

\begin{figure}[t]
\centering
\includegraphics[width=1.0\textwidth]{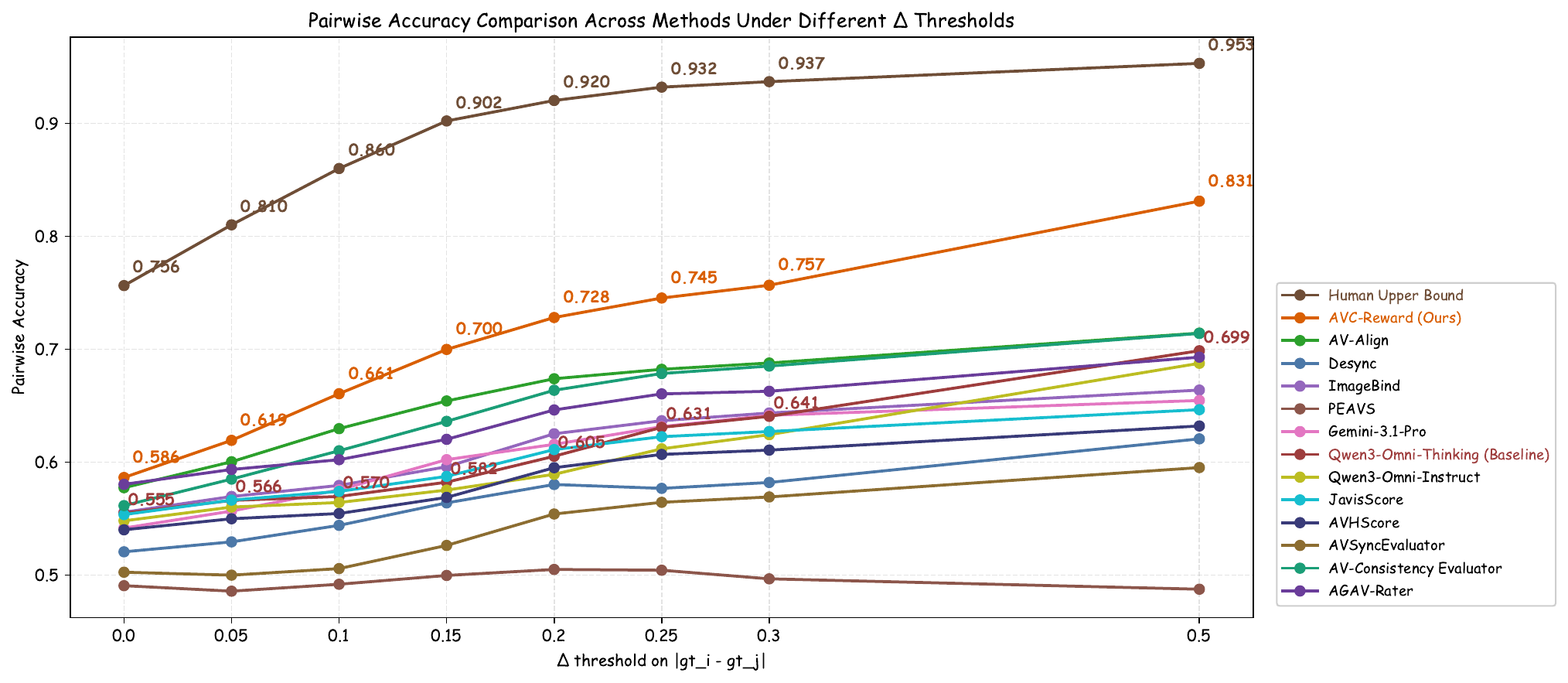}
\caption{Pairwise accuracy under different margin thresholds $\Delta$. Larger $\Delta$ keeps pairs with clearer human-score differences.}

\label{fig:delta_pairwise}
\end{figure}

On AVC-Benchmark, we compare AVC-Reward with the audio-video metrics reviewed in Section~\ref{sec:related_metric} and with zero-shot MLLM judges, and use inter-annotator agreement under each measure as the human upper bound.
The baselines include timing-based metrics (AV-Align~\citep{15avalign} and DeSync~\citep{07desync}), embedding-based metrics (ImageBind~\citep{14imagebind}, JavisScore~\citep{04javisdit}, and AVHScore~\citep{34avhscore}), learned audio-video evaluators (PEAVS~\citep{16peavs}, AVSyncEvaluator~\citep{qian2026beyond}, AV-Consistency Evaluator~\citep{yang2026avbenchhumanalignedautomatedevaluation}, and AGAV-Rater~\citep{pmlr-v267-cao25f}), and zero-shot MLLM judges (Gemini-3.1-Pro~\citep{20gemini3} and Qwen3-Omni~\citep{09qwen3omni}).
For AVC-Reward, we use \(S_{\mathrm{AVC}}\) as its overall prediction, while the human reference follows the overall score defined in App.~\ref{app:benchmark_protocol}.
We report Pairwise, $\Delta$-Pairwise, PLCC (5PL), and SRCC. Since audio-video consistency judgments can be subjective for clips with very similar human scores, $\Delta$-Pairwise computes pairwise accuracy only on pairs whose human-score difference is more than $\Delta$. We use $\Delta=0.2$ as the main setting and report other thresholds in Fig.~\ref{fig:delta_pairwise}.

Overall, AVC-Reward agrees with human judgment better than all automatic baselines. Table~\ref{tab:bench} shows the best result on all four measures. Fig.~\ref{fig:delta_pairwise} gives the same trend across thresholds: as $\Delta$ increases, the human-score differences become clearer and pairwise accuracy improves accordingly. This matches the expected behavior of a human-aligned evaluator: clearer human differences lead to more accurate pairwise judgments. These results support our AVC-Reward as \textbf{a more human-aligned measure of consistency in diverse generated audio-video scenes}.

\subsection{Generation-side Optimization}
\label{sec:exp_gen}

\paragraph{Setup.}
We use OmniNFT~\citep{51omninft}, a strong recent online RL framework for LTX-2~\citep{49ltx2}, to test AVC-Reward in generation optimization. We keep the overall training framework unchanged and replace its original audio-video reward DeSync with AVC-Reward, using its overall score \(S_{\mathrm{AVC}}\) as the optimization signal, with reward-specific learning-rate and weight adjustments for comparable optimization strength. Full training details are provided in App.~\ref{app:generation_rl}.
For evaluation, we avoid JavisBench because part of its prompt set was used to collect training samples, and use AV-Phys-Bench~\citep{cui2026joint}, VABench-T2AV~\citep{17vabench}, and AVGen-Bench~\citep{18avgenbench} as prompt sets. We report the metric set used by OmniNFT on JavisBench and add some audio-video consistency metrics to show changes in this core aspect more clearly.

\newcolumntype{A}{>{\centering\arraybackslash}p{0.68in}}
\newcolumntype{B}{>{\centering\arraybackslash}p{0.54in}}
\newcolumntype{C}{>{\centering\arraybackslash}p{0.52in}}
\newcolumntype{D}{>{\centering\arraybackslash}p{0.58in}}
\newcolumntype{E}{>{\centering\arraybackslash}p{0.52in}}
\newcolumntype{F}{>{\centering\arraybackslash}p{0.50in}}
\newcolumntype{G}{>{\centering\arraybackslash}p{0.52in}}
\newcolumntype{N}{>{\centering\arraybackslash}p{0.53in}}
\definecolor{gainblue}{RGB}{70,96,115}
\newcommand{\gaincell}[1]{\textcolor{gainblue}{#1}}

\begin{table}[t]
\centering
\caption{Generation results across three benchmarks. Gain rows report relative improvements over LTX-2. All other entries are absolute metric values. Best results within each benchmark are in bold.}
\label{tab:gen}
\tablefontsize
\setlength{\tabcolsep}{0.35pt}
\renewcommand{\arraystretch}{0.98}
\begin{tabular*}{\linewidth}{@{}p{1.28in}@{\extracolsep{\fill}}A@{\extracolsep{\fill}}B@{\extracolsep{\fill}}C@{\extracolsep{\fill}}D@{\extracolsep{\fill}}E@{\extracolsep{\fill}}F@{\extracolsep{\fill}}G@{}}
\toprule
\multicolumn{8}{@{}l}{\textbf{(a) AV consistency}} \\
\cmidrule(lr){1-8}
Model & AVC-Score$\uparrow$ & DeSync$\downarrow$ & IB-AV$\uparrow$ & AV-align$\uparrow$ & Javis$\uparrow$ & AVH$\uparrow$ & CAVP$\uparrow$ \\
\specialrule{\lightrulewidth}{0pt}{0pt}
\rowcolor{gray!15}
\multicolumn{8}{@{}c}{\textit{AV-Phys-Bench}} \\
\hline
LTX-2 & 0.756 & 0.376 & 0.234 & 0.158 & 0.189 & 0.226 & 0.812 \\
\arrayrulecolor{gray!65}\hline\arrayrulecolor{black}
OmniNFT (DeSync) & 0.780 & \textbf{0.073} & 0.248 & 0.213 & 0.193 & 0.233 & \textbf{0.819} \\
\gaincell{\hspace*{0.8em}OmniNFT gain} & \gaincell{+3.26\%} & \gaincell{\textbf{+80.60\%}} & \gaincell{+5.84\%} & \gaincell{+34.17\%} & \gaincell{+1.96\%} & \gaincell{+2.91\%} & \gaincell{\textbf{+0.88\%}} \\
\textbf{Ours} (AVC-Reward) & \textbf{0.819} & 0.081 & \textbf{0.288} & \textbf{0.246} & \textbf{0.228} & \textbf{0.276} & 0.818 \\
\gaincell{\hspace*{0.8em}\textbf{Ours gain}} & \gaincell{\textbf{+8.35\%}} & \gaincell{+78.44\%} & \gaincell{\textbf{+22.92\%}} & \gaincell{\textbf{+55.50\%}} & \gaincell{\textbf{+20.47\%}} & \gaincell{\textbf{+22.00\%}} & \gaincell{+0.71\%} \\
\specialrule{\lightrulewidth}{0pt}{0pt}
\rowcolor{gray!15}
\multicolumn{8}{@{}c}{\textit{VABench}} \\
\hline
LTX-2 & 0.764 & 0.421 & 0.216 & 0.185 & 0.177 & 0.207 & 0.805 \\
\arrayrulecolor{gray!65}\hline\arrayrulecolor{black}
OmniNFT (DeSync) & 0.791 & 0.097 & 0.256 & 0.217 & 0.207 & 0.244 & \textbf{0.815} \\
\gaincell{\hspace*{0.8em}OmniNFT gain} & \gaincell{+3.54\%} & \gaincell{+76.97\%} & \gaincell{+18.68\%} & \gaincell{+17.19\%} & \gaincell{+16.65\%} & \gaincell{+18.24\%} & \gaincell{\textbf{+1.31\%}} \\
\textbf{Ours} (AVC-Reward) & \textbf{0.850} & \textbf{0.063} & \textbf{0.277} & \textbf{0.295} & \textbf{0.223} & \textbf{0.262} & 0.814 \\
\gaincell{\hspace*{0.8em}\textbf{Ours gain}} & \gaincell{\textbf{+11.17\%}} & \gaincell{\textbf{+85.03\%}} & \gaincell{\textbf{+28.35\%}} & \gaincell{\textbf{+59.47\%}} & \gaincell{\textbf{+25.65\%}} & \gaincell{\textbf{+26.80\%}} & \gaincell{+1.15\%} \\
\specialrule{\lightrulewidth}{0pt}{0pt}
\rowcolor{gray!15}
\multicolumn{8}{@{}c}{\textit{AVGenBench}} \\
\hline
LTX-2 & 0.786 & 0.288 & 0.257 & 0.246 & 0.189 & 0.231 & 0.803 \\
\arrayrulecolor{gray!65}\hline\arrayrulecolor{black}
OmniNFT (DeSync) & 0.808 & \textbf{0.074} & 0.295 & 0.225 & 0.217 & 0.262 & \textbf{0.816} \\
\gaincell{\hspace*{0.8em}OmniNFT gain} & \gaincell{+2.76\%} & \gaincell{\textbf{+74.31\%}} & \gaincell{+14.86\%} & \gaincell{-8.37\%} & \gaincell{+14.92\%} & \gaincell{+13.20\%} & \gaincell{\textbf{+1.66\%}} \\
\textbf{Ours} (AVC-Reward) & \textbf{0.849} & 0.089 & \textbf{0.310} & \textbf{0.292} & \textbf{0.243} & \textbf{0.286} & 0.814 \\
\gaincell{\hspace*{0.8em}\textbf{Ours gain}} & \gaincell{\textbf{+8.02\%}} & \gaincell{+69.27\%} & \gaincell{\textbf{+20.65\%}} & \gaincell{\textbf{+18.78\%}} & \gaincell{\textbf{+28.74\%}} & \gaincell{\textbf{+23.98\%}} & \gaincell{+1.32\%} \\
\bottomrule
\end{tabular*}

\vspace{0.45em}

\begin{tabular*}{\linewidth}{@{}p{0.60in}*{9}{@{\extracolsep{\fill}}N}@{}}
\toprule
\multicolumn{10}{@{}l}{\textbf{(b) Text alignment and single-modality quality}} \\
\cmidrule(lr){1-10}
Metric & \multicolumn{3}{c}{AV-Phys-Bench} & \multicolumn{3}{c}{VABench} & \multicolumn{3}{c}{AVGenBench} \\
\cmidrule(lr){2-4} \cmidrule(lr){5-7} \cmidrule(lr){8-10}
 & LTX-2 & OmniNFT & \textbf{Ours} & LTX-2 & OmniNFT & \textbf{Ours} & LTX-2 & OmniNFT & \textbf{Ours} \\
\midrule
IB-TV$\uparrow$ & \textbf{0.324} & 0.322 & 0.308 & \textbf{0.248} & 0.242 & 0.232 & \textbf{0.158} & 0.140 & 0.145 \\
IB-TA$\uparrow$ & 0.180 & 0.200 & \textbf{0.207} & 0.127 & \textbf{0.144} & 0.133 & 0.110 & 0.115 & \textbf{0.119} \\
CLIP$\uparrow$ & \textbf{0.287} & 0.282 & 0.272 & \textbf{0.269} & 0.264 & 0.255 & \textbf{0.296} & 0.292 & 0.289 \\
CLAP$\uparrow$ & 0.391 & \textbf{0.402} & 0.399 & 0.363 & \textbf{0.380} & 0.365 & 0.406 & \textbf{0.437} & 0.427 \\
AQ$\uparrow$ & 4.577 & 4.941 & \textbf{4.970} & 4.667 & 5.053 & \textbf{5.087} & 4.882 & 5.154 & \textbf{5.265} \\
VQ$\uparrow$ & 2.471 & 4.262 & \textbf{4.298} & 2.200 & 4.319 & \textbf{4.421} & 2.608 & \textbf{4.157} & 4.129 \\
\bottomrule
\end{tabular*}
\end{table}

\paragraph{Results.}
AVC-Reward consistently improves external audio-video consistency metrics across all prompt sets: IB-AV by $20.65$--$28.35\%$, AV-Align by $18.78$--$59.47\%$, JavisScore by $20.47$--$28.74\%$, and AVHScore by $22.00$--$26.80\%$. Moreover, although OmniNFT directly optimizes DeSync while ours does not, our DeSync gains remain comparable. These gains demonstrate strong generalization to unseen audio-video consistency metrics. CAVP changes by less than $1.7\%$ for both methods, indicating limited separation. As for other evaluate dimensions, because T-A, T-V, and A-V objectives are not perfectly aligned, stronger A-V optimization can slightly reduce text alignment. Within ImageBind, however, the AV gains exceed the observed TA/TV declines, while single-modality quality remains comparable at the OmniNFT level. These results show that \textbf{AVC-Reward is not only an accurate evaluator but can also serve as an effective reward signal for optimizing audio-video consistency in generation}.

\subsection{Human Evaluation}

\begin{figure}[t]
\centering
\includegraphics[width=1.0\linewidth]{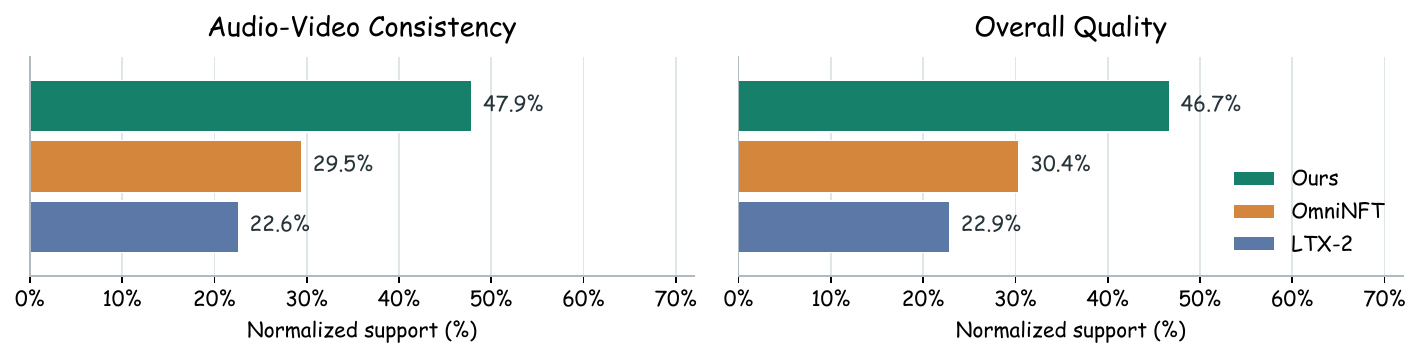}
\caption{Overall human preference across the three benchmarks.}
\label{fig:human_eval_overall}
\end{figure}

\label{sec:exp_human}
We randomly sample 100 prompts from AVGenBench, AV-Phys-Bench, and VABench, and compare LTX-2, OmniNFT, and Ours with blinded randomized human judgments, as summarized in Fig.~\ref{fig:human_eval_overall}.
We ask 10 annotators with undergraduate degrees or higher to evaluate the samples.
Ours receives the highest preference for both audio-video consistency and overall quality. These preferences further \textbf{demonstrate the effectiveness of our generation-side optimization from a human perceptual perspective}. detailed rules and benchmark-wise results are reported in App.~\ref{app:human_eval}.

\subsection{Case Studies}
\label{sec:exp_case}

\begin{figure}[!t]
\centering
\includegraphics[width=1.0\textwidth]{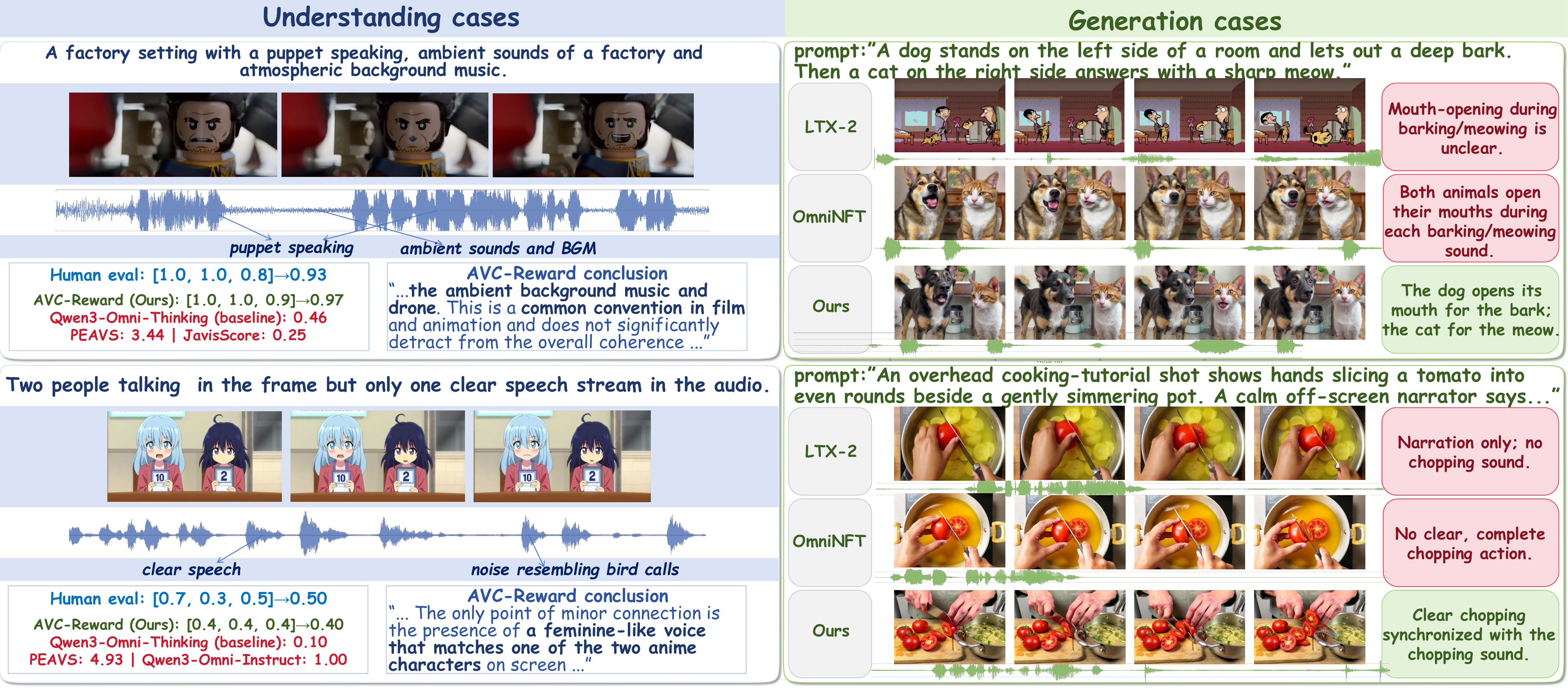}
\caption{Representative cases. Left: AVC-Reward produces scores closer to human judgments and identifies specific consistency issues. Right: generation results of LTX-2, OmniNFT, and Ours.}
\label{fig:cases}
\end{figure}

Fig.~\ref{fig:cases} presents representative cases of AVC-Reward in evaluation and generation optimization. AVC-Reward handles multi-layer audio, diverse relations such as off-screen and ambient sounds, and multiple or concurrent events by distinguishing audio roles and evaluating event relations separately; as an RL reward, it also improves generation under these complexities. Moreover, bidirectional cross-modal verification enables finer checks of sound attributes and visible sound-producing actions, while the optimized generator produces more accurate sounds and clearer, more correct corresponding visual evidence. More cases are provided in App.~\ref{app:case}.

\begin{table}[t]

\centering

\caption{Ablations and aggregation analysis on AVC-Benchmark.}

\label{tab:ablation}

\tablefontsize

\setlength{\tabcolsep}{3.0pt}

\renewcommand{\arraystretch}{0.98}

\begin{tabular}{@{}lccc@{\hspace{1.0em}}lccc@{}}

\toprule

\multicolumn{4}{c}{\textbf{Training Strategy}} & 
\multicolumn{4}{c}{\textbf{Score Aggregation}} \\

\cmidrule(lr){1-4}\cmidrule(l){5-8}

Variant & $\Delta$-Pair$\uparrow$ & PLCC$\uparrow$ & SRCC$\uparrow$ &
Variant & $\Delta$-Pair$\uparrow$ & PLCC$\uparrow$ & SRCC$\uparrow$ \\

\hline

Base & 0.605 & 0.347 & 0.154 &
$S_{\mathrm{mac}}$ only & 0.688 & 0.467 & \textbf{0.251} \\

GRPO only & 0.630 & 0.330 & 0.161 &
$S_{\mathrm{temp}}$ only & 0.717 & 0.479 & 0.235 \\

SFT only & 0.679 & 0.295 & \textbf{0.244} &
$S_{\mathrm{attr}}$ only & 0.715 & 0.482 & 0.240 \\ \hline

\textbf{SFT+GRPO} & \textbf{0.728} & \textbf{0.494} & \textbf{0.244} &
\textbf{Mean ($S_{\mathrm{AVC}}$)} & \textbf{0.728} & \textbf{0.494} & 0.244 \\

\bottomrule

\end{tabular}

\end{table}

\subsection{Training Ablation and Score-aggregation Analysis}
\label{sec:exp_ablation}

We examine the training strategy and score aggregation in Table~\ref{tab:ablation}.
Compared with GRPO only, SFT+GRPO achieves better human alignment, demonstrating the effectiveness of the proposed five-stage reasoning supervision, while the improvement from SFT only shows the benefit of human preference alignment through GRPO.
For score aggregation, averaging all three dimensions achieves higher $\Delta$-Pairwise than using single dimension alone, supporting their complementary use in \(S_{\mathrm{AVC}}\).

\section{Conclusion}
\label{sec:conclusion}
We study general audio-video consistency evaluation under the challenges of \textbf{multi-event composition}, \textbf{multi-layer audio mixture}, and \textbf{asymmetric cross-modal relations}. We propose an \textbf{event-centric coarse-to-fine three-level metric} and instantiate it as \textbf{AVC-Reward}, trained with structured SFT and human-aligned GRPO. We further introduce AVC-Benchmark to evaluate human agreement across diverse audio-video generation models. AVC-Reward achieves the strongest human alignment among evaluated methods and, when integrated into OmniNFT, effectively improves audio-video consistency. These results demonstrate the effectiveness of event-centric consistency evaluation as both a reliable metric and an optimization signal for joint audio-video generation.

\bibliography{iclr2027_conference}
\bibliographystyle{iclr2027_conference}

\clearpage
\appendix

\appendix
\section*{Appendix}
\addcontentsline{toc}{section}{Appendix}

In the Appendix, we present the following details:

\begin{itemize}
    \item Complete formal metric definitions in Section~\ref{app:metric}.
    \item Data construction and annotation details in Section~\ref{app:pipeline}.
    \item Training details in Section~\ref{app:training_details}.
    \item AVC-Benchmark details in Section~\ref{app:benchmark}.
    \item Additional evaluation and generation case studies in Section~\ref{app:case}.
    \item Human evaluation protocol and results in Section~\ref{app:human_eval}.
\end{itemize}

\section{Formal Metric Definitions}
\label{app:metric}

This section gives the complete mathematical form of the event-centric protocol in Section~\ref{sec:method}.  The protocol evaluates an observed audio--video pair from coarse to fine: it first compares the two global descriptions, then verifies event existence and timing, and finally checks event attributes.  The three scores remain separate throughout the evaluation; when one scalar is needed, we use their arithmetic mean.

\subsection{Clip and modality representations}
\label{app:metric_representation}

Let an observed clip be
\begin{equation}
    \mathcal{P}=(\mathcal{A},\mathcal{V}), \qquad
    \mathcal{T}=[0,T],
\end{equation}
where $\mathcal{A}$ and $\mathcal{V}$ denote the complete audio and video streams on the shared time interval $\mathcal{T}$.  The protocol evaluates the observed pair directly, so the provenance of either stream is irrelevant.  The first two stages convert the streams into
\begin{equation}
    \mathcal{H}^{A}=(\mathcal{G}^{A},\mathcal{E}^{A}), \qquad
    \mathcal{H}^{V}=(\mathcal{G}^{V},\mathcal{E}^{V}),
    \label{eq:app_modal_repr}
\end{equation}
where $\mathcal{G}^{A}$ and $\mathcal{G}^{V}$ are full-clip global scene and narrative descriptions, and $\mathcal{E}^{A}$ and $\mathcal{E}^{V}$ are event sets.  Global descriptions may include the scene or environment, entities and their roles, and the affective or narrative tone.  They summarize the whole clip across $\mathcal{T}$.

An audio event is represented as
\begin{equation}
 e_i^{A}=\bigl(t_{i,s}^{A},t_{i,e}^{A},c_i^{A},\lambda_i^{A},\phi_i^{A}\bigr),
 \qquad 0\leq t_{i,s}^{A}\leq t_{i,e}^{A}\leq T,
 \label{eq:app_audio_event}
\end{equation}
where $c_i^{A}$ is the event category, $\lambda_i^{A}$ is its source role, and $\phi_i^{A}$ contains the observable acoustic attributes.  The categories used by the annotation protocol are human speech, music or rhythm, and other physical sound events.  The source role is
\begin{equation}
    \lambda_i^{A}\in\{\mathrm{physical},\mathrm{narrative}\}.
\end{equation}
Physical events belong to the recorded world.  Narrative events, such as background music or narration added for storytelling, are judged by their relation to the scene; visible source evidence is optional.  For a physical event, the annotation also records whether the source is on screen or off screen when this distinction is observable.

A visual event is represented as
\begin{equation}
 e_j^{V}=\bigl(t_{j,s}^{V},t_{j,e}^{V},c_j^{V},\phi_j^{V}\bigr),
 \qquad 0\leq t_{j,s}^{V}\leq t_{j,e}^{V}\leq T,
 \label{eq:app_visual_event}
\end{equation}
where $c_j^{V}$ identifies the audio-producing action (for example, speaking, playing an instrument, an object interaction, or an environmental action) and $\phi_j^{V}$ records its observable morphology, source location, and motion.  Silent actions are retained as context; they create no required audio event.

\subsection{Five-stage evaluation protocol}
\label{app:metric_protocol}

The five stages are formal interfaces between the evidence records and the final score.  Stages 1 and 2 analyze one modality at a time:
\begin{align}
    \mathcal{H}^{A} &= F_A(\mathcal{A}),
    &\mathcal{H}^{V} &= F_V(\mathcal{V}).
    \label{eq:app_stage12}
\end{align}
Stage 1 identifies the acoustic layers, global audio context, speaker or music information, and time-localized audio events.  Stage 2 identifies the visual scene, people and actions, and the time-localized visual events that can produce sound.

Stages 3 and 4 verify the two directions separately:
\begin{equation}
    \mathcal{R}^{A\rightarrow V}
    =F_{A\rightarrow V}(\mathcal{H}^{A},\mathcal{V}),
    \qquad
    \mathcal{R}^{V\rightarrow A}
    =F_{V\rightarrow A}(\mathcal{H}^{V},\mathcal{A}).
    \label{eq:app_stage34}
\end{equation}
For every audio event, Stage 3 checks whether the video supplies the required evidence in the event's time interval.  For a short event this is a direct check of the relevant action.  For a continuous event it checks the minimal visible evidence throughout the interval.  For every visual event, Stage 4 first decides whether the action should produce sound.  Missing audio for a naturally silent action is reasonable silence; missing audio for a sound-producing action is an existence failure.  Events that pass the existence judgment proceed to physical or attribute checks.

The output of either verification stage is a relation record
\begin{equation}
    r_k=(e_i^{A},e_j^{V},\rho_k),
    \qquad
    e_i^{A}\in\mathcal{E}^{A}\cup\{\varnothing\},\quad
    e_j^{V}\in\mathcal{E}^{V}\cup\{\varnothing\},
    \label{eq:app_relation}
\end{equation}
where $\rho_k$ specifies the relation being checked and stores the corresponding evidence.  Either event can be absent because the relation is asymmetric or because the expected cross-modal evidence is missing.  We write
\begin{equation}
    \mathcal{R}=\mathcal{R}^{A\rightarrow V}\cup\mathcal{R}^{V\rightarrow A}.
    \label{eq:app_all_relations}
\end{equation}

Stage 5 uses the global descriptions and the verified relations to produce
\begin{equation}
    \mathcal{S}
    =\bigl(S_{\mathrm{mac}},S_{\mathrm{temp}},S_{\mathrm{attr}}\bigr)
    \in[0,1]^3,
    \label{eq:app_score_tuple}
\end{equation}
with
\begin{equation}
    S_{\mathrm{mac}}=\Phi_{\mathrm{mac}}(\mathcal{G}^{A},\mathcal{G}^{V}),
    \qquad
    S_{\mathrm{temp}}=\Phi_{\mathrm{temp}}(\mathcal{R}),
    \qquad
    S_{\mathrm{attr}}=\Phi_{\mathrm{attr}}(\mathcal{R}).
    \label{eq:app_score_functions}
\end{equation}

\subsection{Three-level consistency scores}
\label{app:metric_scores}

\paragraph{Global narrative harmony.}
$S_{\mathrm{mac}}$ compares the audio and video global descriptions over the entire interval $\mathcal{T}$.  The comparison covers narrative content and entity identity, scene or environment, and affective tone.  A persistent conflict in any of these aspects lowers the score even when individual events happen to overlap in time.  A voice-over or background music can be consistent without a visible source when its scene, narrative role, and affective relation are plausible.

\paragraph{Event-level temporal synchronization.}
For each relation $r_k$, let $x_k\in\{0,1\}$ denote the existence judgment and let $y_k\in[0,1]$ denote the onset--offset alignment judgment.  The existence judgment is defined by the source role:
\begin{equation}
 x_k=
 \begin{cases}
  1, & \text{bidirectional evidence supports an on-screen physical event},\\
  1, & \text{an off-screen physical event is plausible and its consequences are consistent},\\
  1, & \text{a narrative event has an appropriate semantic or rhythmic role},\\
  1, & \text{the visual action is naturally silent (reasonable silence)},\\
  0, & \text{otherwise.}
 \end{cases}
 \label{eq:app_existence}
\end{equation}
For an on-screen physical event, bidirectional evidence means both directions are checked: a visible sound-producing action must have corresponding audio, and an audible on-screen source must have corresponding visual evidence.  Each relation record contains an existence judgment $x_k\in\{0,1\}$ and a temporal-alignment judgment $y_k\in[0,1]$.  The temporal judgment covers event onset, offset, duration, and the alignment of relevant states for rhythmic or continuous events.  Narrative events use a meaningful rhythmic or temporal relation; reasonable-silence records receive full temporal evidence.  The final LLM judge reads all temporal evidence in $\mathcal{R}$ and outputs
\begin{equation}
    S_{\mathrm{temp}}=\Phi_{\mathrm{temp}}(\mathcal{R})\in[0,1].
    \label{eq:app_stemp}
\end{equation}
The temporal score covers event existence and timing.  Material, timbre, mouth shape, finger motion, and reaction plausibility belong to the attribute evidence.

\paragraph{Fine-grained attribute alignment.}
Attribute evaluation proceeds whenever temporal correspondence establishes a tentative audio--visual event pair within an acceptable tolerance. The verifier records applicable physical, morphological, and causal-response evidence for each relation.  Physical evidence covers material or timbre, energy or loudness, and spatial or distance cues.  Morphological evidence covers fine visual states such as mouth shape, finger motion, and the visible action that produces the sound.  Causal-response evidence covers source localization, emotional response, and response magnitude for off-screen or narrative evidence.  The final LLM judge reads all applicable attribute evidence in $\mathcal{R}$ and outputs
\begin{equation}
    S_{\mathrm{attr}}=\Phi_{\mathrm{attr}}(\mathcal{R})\in[0,1].
    \label{eq:app_sattr}
\end{equation}

\subsection{Overall AVC score}
\label{app:metric_aggregation}

The final judgment follows the order $S_{\mathrm{mac}}\rightarrow S_{\mathrm{temp}}\rightarrow S_{\mathrm{attr}}$.  The three outputs remain separate diagnostic dimensions.  When an overall score is required, we use the arithmetic mean
\begin{equation}
    S_{\mathrm{AVC}}
    =\frac{1}{3}\left(S_{\mathrm{mac}}+S_{\mathrm{temp}}+S_{\mathrm{attr}}\right).
    \label{eq:app_savc}
\end{equation}
The overall score is the unweighted arithmetic mean of the three separately reported dimensions.
\raggedbottom
\section{Data Construction and Annotation}
\label{app:taxonomy}
\label{app:pipeline}

\subsection{Training Data Sources and Statistics}
\label{sec:app_data_source}

We construct two data pools for AVC-Reward training, one for structured-reasoning SFT and the other for human calibration. Table~\ref{tab:training_data_sources} summarizes the sources and statistics of the training data used for SFT and GRPO.
Throughout this process, a wide array of multimodal models~\citep{09qwen3omni, 13ovi, 44mova, pan2025generative, pan2026janus, tu2024motioneditor, tu2025stableavatar, tu2025stableanimator, 25hunyuanfoley, 14imagebind, 26infinitetalk, 04javisdit} is leveraged to curate more comprehensive data along with precise annotations.

\paragraph{SFT source pool.}
We first collect ground-truth audio-video clips from two categories: general scenes and human-centered scenes. We use Qwen3-Omni~\citep{09qwen3omni} to filter the clips for basic audio-video quality and clarity, salient event and temporal relations, and informative macro-semantic content, retaining 2.5k general clips and 1k human-centered clips. The four filtering questions are provided in our project's GitHub repository. Qwen3-Omni then generates audio and video captions for the retained clips, which are used to construct samples under different generation paradigms.

For the general subset, Ovi~\citep{13ovi} generates 2.5k JAVG samples from full audio-video captions, while HunyuanVideo-Foley~\citep{25hunyuanfoley} generates 2.5k V2A samples from the ground-truth videos and corresponding audio captions. For the human-centered subset, Ovi generates 1k JAVG samples, while InfiniteTalk~\citep{26infinitetalk} and an internal model each generate 1k A2V samples using the ground-truth audio with visual conditioning. Together with the 3.5k ground-truth clips, this gives 11.5k candidate samples. We further filter the generated pool for basic audio-video quality and annotation completeness, retaining 8k clips for SFT.

\paragraph{Human-calibration source pool.}
We construct the human-calibration source pool from JavisBench~\citep{04javisdit}. Its relatively large scale, paired ground-truth references, and broad content coverage provide a practical starting point for constructing diverse positive and negative examples when high-quality audio-video data remain limited.
We sample 3.2k clips from each of five sources: JavisBench ground truth, Ovi-I2V, Ovi-T2V, HunyuanVideo-Foley, and an internal A2V model. The generated samples use the corresponding JavisBench inputs, covering JAVG, V2A, and A2V paradigms. These 16k clips form the initial candidate pool. We run the SFT model $f_1$ eight times on each candidate to estimate sample difficulty and prioritize cases for which the model has not yet formed stable judgments. To preserve content diversity, we cluster ImageBind~\citep{14imagebind} audio-video embeddings into 300 clusters and sample across clusters. We also construct global-narrative negatives by swapping audio tracks across clusters. Finally, we sample original candidates and constructed global negatives at a 3:1 ratio to obtain the 10k rollout set used for GRPO.

\begin{table}[t]
\centering
\small
\setlength{\tabcolsep}{4pt}
\begin{tabular}{lllrcc}
\toprule
\textbf{Subset} & \textbf{Source model} & \textbf{Paradigm} & \textbf{Raw clips} & \textbf{SFT} & \textbf{Human calib.} \\
\midrule
General & Ground truth & GT & 2.5k & \checkmark & \\
General & Ovi & JAVG & 2.5k & \checkmark & \\
General & HunyuanVideo-Foley & V2A & 2.5k & \checkmark & \\
Human & Ground truth & GT & 1.0k & \checkmark & \\
Human & Ovi & JAVG & 1.0k & \checkmark & \\
Human & InfiniteTalk & A2V & 1.0k & \checkmark & \\
Human & Internal model & A2V & 1.0k & \checkmark & \\
\midrule
JavisBench & Ground truth & GT & 3.2k & & \checkmark \\
JavisBench & Ovi-I2V & JAVG & 3.2k & & \checkmark \\
JavisBench & Ovi-T2V & JAVG & 3.2k & & \checkmark \\
JavisBench & HunyuanVideo-Foley & V2A & 3.2k & & \checkmark \\
JavisBench & Internal model & A2V & 3.2k & & \checkmark \\
\bottomrule
\end{tabular}
\caption{Sources and statistics of the training data. The SFT source pool contains 11.5k candidate clips, of which 8k are retained after supervision filtering. The human-calibration source pool contains 16k candidates, from which the 10k GRPO rollout set is constructed.}
\label{tab:training_data_sources}
\end{table}

The resulting data pool covers different generation paradigms and provides the source samples for the two complementary supervision pipelines described below.

\subsection{Five-Stage Automated Annotation Pipeline}
\label{sec:app_annotation_pipeline}

\begin{figure}[t]
    \centering
    \includegraphics[width=\linewidth]{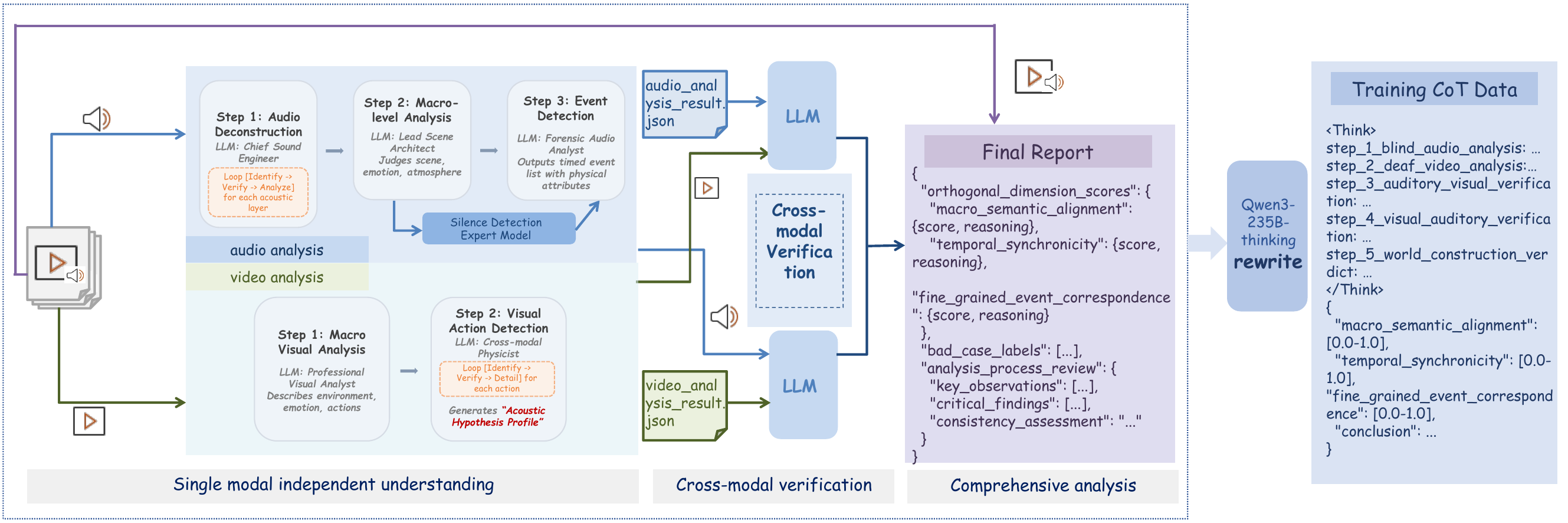}
    \caption{
    Overview of the five-stage automated annotation pipeline.
    Audio and video are first analyzed independently, followed by bidirectional
    event verification and final aggregation into three consistency scores
    and a diagnostic conclusion.
    }
    \label{fig:auto_annotation_pipeline}
\end{figure}

Informed by practical insights of prior approaches~\citep{geng2025longvale, zhou2026av, leng2026preference, huang2026va, pan2026omniweaving, yang2026arcflow, sun2024auto, mei2024wavcaps}, To construct structured supervision at scale, we implement a five-stage automated annotation pipeline that first analyzes audio and video independently, then verifies event relations in both directions, and finally aggregates the structured evidence into the three consistency levels.
Table~\ref{tab:auto_annotation_pipeline} summarizes the input, operation, tool, and output of each stage.
Fig.~\ref{fig:auto_annotation_pipeline} illustrates the complete data flow.

\begin{table}[t]
\centering
\footnotesize
\setlength{\tabcolsep}{3pt}
\renewcommand{\arraystretch}{1.12}
\begin{tabular}{
 >{\raggedright\arraybackslash}p{0.12\linewidth}
 >{\raggedright\arraybackslash}p{0.15\linewidth}
 >{\raggedright\arraybackslash}p{0.25\linewidth}
 >{\raggedright\arraybackslash}p{0.15\linewidth}
 >{\raggedright\arraybackslash}p{0.23\linewidth}}
\toprule
\textbf{Stage} &
\textbf{Input} &
\textbf{Operation} &
\textbf{Tool} &
\textbf{Output} \\
\midrule

1. Audio analysis &
Audio $A$ + SED hints $T_{\mathrm{SED}}$ &
Analyze the audio scene, layers, events, and timestamps. &
Gemini 2.5 Pro + SED hints &
$H_A=(G_A,E_A)$: audio context and timestamped audio events. \\

2. Video analysis &
Video $V$ &
Analyze the visual scene and localize audio-relevant events. &
Qwen3-VL-Max &
$H_V=(G_V,E_V)$: visual context and timestamped visual events. \\

3. A$\rightarrow$V verification &
$H_A$ from Stage 1 + $V$ &
Verify visual evidence, timing, and applicable event attributes for each audio event. &
Qwen3-VL-Max &
$R_{A\rightarrow V}$: audio-to-video verification records. \\

4. V$\rightarrow$A verification &
$H_V$ from Stage 2 + $A$ &
Verify expected audio, reasonable silence, timing, and applicable event attributes for each visual event. &
Gemini 2.5 Pro &
$R_{V\rightarrow A}$: video-to-audio verification records. \\

5. Aggregation &
$\begin{gathered}H_A,H_V,\\R_{A\rightarrow V},R_{V\rightarrow A}\end{gathered}$ + original AV &
Re-examine the clip and aggregate the verified evidence. &
Gemini 2.5 Pro &
$(S_{\mathrm{mac}},S_{\mathrm{temp}},S_{\mathrm{attr}},\kappa)$. \\

\bottomrule
\end{tabular}
\caption{
Summary of the five-stage automated annotation pipeline.
Stages 1--2 produce modality-specific analysis records, Stages 3--4 verify their event relations in the opposite modality, and Stage 5 aggregates all records into the final three-level judgment.
}
\label{tab:auto_annotation_pipeline}
\end{table}

\paragraph{Pipeline details.}
Table~\ref{tab:auto_annotation_pipeline} and Fig.~\ref{fig:auto_annotation_pipeline} summarize the complete annotation procedure.

Stages 1 and 2 first parse the two modalities independently.
The audio branch separates speech, music, and other sound events, and records their scene context, speaker information, acoustic attributes, and timestamps.
The video branch extracts the global scene, people, emotions, and timestamped visual events that may be relevant to audio.

Stages 3 and 4 then perform bidirectional event verification.
For each audio event, audio-to-video verification searches its temporal window for supporting visual evidence and, when applicable, examines temporal alignment and fine-grained physical attributes.
The absence of a visible source is retained as evidence rather than directly treated as an inconsistency, allowing later reasoning over off-screen audio and other asymmetric relations.
Conversely, video-to-audio verification first determines whether a visual event should naturally produce sound.
Naturally silent actions are preserved as reasonable silence, while sound-producing actions are checked for corresponding acoustic evidence, temporal alignment, and physical compatibility.

Stage 5 aggregates the modality-specific analyses and the two directional verification records.
The evaluator also re-examines the original audio-video clip instead of relying only on intermediate annotations.
It then produces scores for global narrative harmony, event-level temporal synchronization, and fine-grained attribute alignment, together with a diagnostic conclusion that summarizes the main consistent or inconsistent events.

\paragraph{Harness-based process control.}
The three evaluation levels require several non-trivial judgments, so we do not rely on a single unconstrained model response to construct the supervision.
Instead, the annotation harness introduces explicit process controls around the underlying expert models.
First, audio and video are analyzed independently, preventing early cross-modal information from directly biasing event discovery.
Second, each detected event is converted into explicit timestamps and evidence records, and later checks operate on these records rather than restarting from an unconstrained global judgment.
Third, verification is conditional: existence is checked before temporal or attribute compatibility, and inapplicable cases such as reasonable silence are explicitly bypassed.
Fourth, the same relation is checked in both audio-to-video and video-to-audio directions, providing a cross-check for missing and hallucinated events.
Finally, the aggregation stage receives all intermediate evidence but also re-examines the original audio-video clip, forming a final feedback check against errors propagated from earlier stages.

These controlled verification records are then transferred to AVC-Reward through both training stages.
During SFT, the complete five-stage annotation trajectory, including single-modal analyses, bidirectional event verification, three scores, and the diagnostic conclusion, is used as the full supervision target.
The model is therefore trained to reproduce the same staged verification process rather than learning only a direct mapping from the input clip to final scores.
During GRPO, the structure reward explicitly rewards the presence of all five reasoning stages, while score distillation aligns the resulting judgments with human-calibrated scores.
Together, the two stages preserve the verification procedure established by the annotation harness while transferring it from the cascaded expert pipeline into a single reward model.
Whether this learned procedure leads to correct judgments is further evaluated through human alignment, training ablations, and qualitative cases.

\subsection{Human Annotation and Score Calibration}
\label{sec:app_human_annotation}

In human annotation, we found that score-only judgments showed noticeable subjective variation and made disagreements difficult to verify.
We therefore design a 39-label taxonomy that asks annotators to record explicit consistency states and failure types before score calibration, reducing unconstrained subjectivity while preserving observable information about event presence, timing, audio roles, and cross-modal relations.
The taxonomy follows the same three dimensions as AVC-Reward: global labels cover narrative, affective, and environmental consistency; temporal labels separately describe physical interactions, speaker events, and music/rhythmic events, including missing, hallucinated, off-screen, and narrative-layer cases; fine-grained labels describe material, intensity, spatial, motion, and causal-response attributes.
Table~\ref{tab:human_taxonomy} summarizes the complete annotation space.

\begin{table*}[t]
\centering
\small
\setlength{\tabcolsep}{4pt}
\renewcommand{\arraystretch}{1.12}
\begin{tabular}{
 >{\raggedright\arraybackslash}p{0.17\textwidth}
 >{\raggedright\arraybackslash}p{0.12\textwidth}
 >{\raggedright\arraybackslash}p{0.09\textwidth}
 >{\raggedright\arraybackslash}p{0.52\textwidth}}
\toprule
\textbf{Dimension} &
\textbf{Event type} &
\textbf{Label} &
\textbf{Annotation meaning} \\
\midrule

\multirow{5}{*}{\parbox{0.17\textwidth}{\raggedright Global narrative harmony}}
& \multirow{5}{*}{Global}
& M-A & Globally consistent \\
& & M-B & Narrative-content conflict \\
& & M-C & Emotion or atmosphere mismatch \\
& & M-D & Ambient sound or scene-setting conflict \\
& & M-E & Other global semantic inconsistency \\
\midrule

\multirow{25}{*}{\parbox{0.17\textwidth}{\raggedright Event-level temporal synchronization}}

& \multirow{8}{*}{\parbox{0.12\textwidth}{\raggedright Physical interaction}}
& T-P-A & On-screen event correctly matched \\
& & T-P-B & On-screen event with missing audio \\
& & T-P-C & Audio hallucination or implausible off-screen audio \\
& & T-P-D & Temporal offset between action and sound \\
& & T-P-E & Plausible off-screen audio \\
& & T-P-F & No applicable event \\
& & T-P-G & Other temporal mismatch \\
& & T-P-H & Unjudgeable due to severe global inconsistency \\
\cmidrule(lr){2-4}

& \multirow{8}{*}{Speaker}
& T-S-A & Physical-layer speech correctly matched \\
& & T-S-B & Missing speech audio \\
& & T-S-C & Speech audio hallucination \\
& & T-S-D & Speech-stream onset/offset mismatch \\
& & T-S-E & Narrative-layer voice-over \\
& & T-S-F & No applicable speaker event \\
& & T-S-G & Other speaker temporal mismatch \\
& & T-S-H & Unjudgeable due to severe global inconsistency \\
\cmidrule(lr){2-4}

& \multirow{9}{*}{\parbox{0.12\textwidth}{\raggedright Music / rhythmic}}
& T-M-A & Physical-layer music correctly matched \\
& & T-M-B & Missing music audio \\
& & T-M-C & Music audio hallucination \\
& & T-M-D & Physical-layer rhythmic misalignment \\
& & T-M-E & Appropriate narrative-layer background music \\
& & T-M-F & Narrative-layer rhythmic misalignment \\
& & T-M-G & No applicable music or rhythmic event \\
& & T-M-H & Other music or rhythmic temporal mismatch \\
& & T-M-I & Unjudgeable due to severe global inconsistency \\
\midrule

\multirow{9}{*}{\parbox{0.17\textwidth}{\raggedright Fine-grained attribute alignment}}
& \multirow{9}{0.12\textwidth}{\raggedright Event attribute}
& F-A & Fully aligned \\
& & F-B & Material or timbre mismatch \\
& & F-C & Energy or intensity mismatch \\
& & F-D & Spatial or distance mismatch \\
& & F-E & Fine-grained state or motion mismatch \\
& & F-F & Implausible causal reaction \\
& & F-G & Event does not pass temporal verification \\
& & F-H & Other fine-grained attribute mismatch \\
& & F-I & Unjudgeable due to severe global inconsistency \\

\bottomrule
\end{tabular}
\caption{
Human annotation taxonomy with 39 labels.
Global narrative harmony evaluates clip-level narrative and scene coherence; event-level temporal synchronization separately covers physical interactions, speaker events, and music or rhythmic events; fine-grained attribute alignment examines the internal audio-visual attributes of temporally matched events.
}
\label{tab:human_taxonomy}
\end{table*}

\paragraph{Annotation protocol.}
Annotators judge the three dimensions independently.
Within each dimension, all applicable failure labels can be selected, allowing a clip to contain multiple inconsistency types or multiple affected events.
For temporal synchronization, annotators additionally distinguish physical-layer on-screen events, plausible off-screen events, and narrative-layer audio such as voice-over and background music.
A severe global conflict can be marked as unjudgeable for a later event-level category when the corresponding relation cannot be meaningfully assessed, but this does not otherwise propagate the global judgment to the other dimensions.

For each annotated sample, three annotators independently provide the dimension labels.
The training and test sets are strictly disjoint.
The held-out test set is additionally assigned to annotators who do not participate in labeling the training set, preventing overlap in both samples and annotators between model training and final human evaluation.

\paragraph{Label-to-score calibration and score rewriting validation.}

Human annotations provide detailed categorical supervision, while AVC-Reward requires continuous targets for the three dimensions. Directly collecting continuous scores for all training samples introduces annotation subjectivity and variance, making large-scale score supervision difficult to verify and maintain consistently. Therefore, we investigate whether human-verified labels can serve as an intermediate representation for recovering continuous scores.

We first study the relationship between human-verified labels and continuous scores on AVC-Benchmark, where both label annotations and human score annotations are available. We analyze whether label configurations consistently correspond to similar human scores. Pairs sharing the same human-verified label configuration have an average absolute score difference of only 0.0321 (median 0.0111), compared with 0.2002 (median 0.1000) for pairs with different configurations (Table~\ref{tab:label_score_consistency}). This indicates that label configurations strongly constrain human scores and motivates label-to-score calibration for continuous reward construction.

\begin{table}[t]
\centering
\small
\setlength{\tabcolsep}{5pt}
\begin{tabular}{lrrr}
\toprule
\textbf{Label configuration} &
\textbf{Pairs} &
\textbf{Mean diff.} &
\textbf{Median diff.} \\
\midrule
Same      & 6,717   & \textbf{0.0321} & \textbf{0.0111} \\
Different & 242,148 & 0.2002          & 0.1000 \\
\bottomrule
\end{tabular}
\caption{
Absolute differences between human scores for sample pairs with the same or different label configurations.
}
\label{tab:label_score_consistency}
\end{table}

Based on this observation, we further investigate whether a large language model can learn the label-to-score mapping through few-shot calibration. Specifically, we extract 10 representative label-to-score mappings from the AVC-Benchmark annotations and provide them to Qwen3-235B as few-shot demonstrations for score rewriting.

Given these demonstrations, Qwen3-235B maps each label record to $(S_{\mathrm{mac}},S_{\mathrm{temp}},S_{\mathrm{attr}})$ and a diagnostic conclusion. The prompt specifies label semantics, failure severity, multi-label handling, and output format; details are provided in our project's GitHub repository.

We evaluate the rewriting procedure on AVC-Benchmark samples with human scores, excluding the manually selected few-shot examples used for prompt construction. As shown in Table~\ref{tab:rewrite_validation}, the rewritten scores preserve human ordering across all three dimensions; the averaged score achieves PLCC 0.780 and $\Delta$-Pairwise@0.2 of 0.905.

\begin{table}[t]
\centering
\small
\setlength{\tabcolsep}{6pt}
\begin{tabular}{lrrr}
\toprule
\textbf{Score} &
\textbf{PLCC} &
\textbf{$\Delta$-Pairwise@0.2} \\
\midrule
Global narrative harmony          & 0.591 & 0.712 \\
Temporal synchronization          & 0.454 & 0.816 \\
Fine-grained attribute alignment  & 0.601 & 0.798 \\
Overall                           &\textbf{0.780} & \textbf{0.905} \\
\bottomrule
\end{tabular}
\caption{
Agreement between rewritten and human scores after excluding the manually selected few-shot examples.
}
\label{tab:rewrite_validation}
\end{table}

\begin{table}[t]
\centering
\small
\setlength{\tabcolsep}{6pt}
\begin{tabular}{lrrr}
\toprule
\textbf{Setting} &
\textbf{Comparisons} &
\textbf{$r_{\mathrm{distill}}$} &
\textbf{$1-r_{\mathrm{distill}}$} \\
\midrule
Rewrite, full evaluation set & 706   & 0.853 & 0.147 \\
Human leave-one-out          & 2,118 & 0.881 & 0.119 \\
\bottomrule
\end{tabular}
\caption{
Score-rewriting agreement measured with the score-distillation reward used during GRPO. Human leave-one-out results provide a reference for annotator variation.
}
\label{tab:rewrite_rdistill}
\end{table}
We further evaluate score rewriting in the score-distillation reward space used for GRPO (Table~\ref{tab:rewrite_rdistill}). Rewritten scores achieve $r_{\mathrm{distill}}=0.853$, compared with 0.881 for human leave-one-annotator-out references. The corresponding reward-space discrepancies, measured by $1-r_{\mathrm{distill}}$, are 0.147 and 0.119, respectively. These results suggest that rewriting errors are comparable in magnitude to the observed inter-annotator variability, with a modest additional discrepancy of 0.028 in the reward space directly relevant to GRPO supervision.

The similar rewriting performance on the full and evaluation sets indicates that the few-shot label-to-score mapping generalizes beyond the manually selected examples. The remaining gap to human leave-one-out agreement reflects the approximation introduced by automatic score rewriting.

For GRPO training, we construct a separate label-to-score calibration prompt using only the GRPO training data. Specifically, we randomly sample 100 examples from the GRPO data pool and collect additional continuous score annotations from human annotators. From these examples, we extract another 10 representative label-to-score mappings and use them as few-shot demonstrations for Qwen3-235B to rewrite scores for GRPO training samples. This calibration process is completely isolated from AVC-Benchmark, and no benchmark annotations are used for GRPO score rewriting or training.

Our project's GitHub repository provides the complete rewriting prompt.

\flushbottom
\section{Training Details}
\label{app:training_details}

\subsection{SFT Cold Start}
\label{app:sft}

We initialize AVC-Reward from Qwen3-Omni-Thinking and perform full-parameter fine-tuning using the 8k structured annotation samples described in Sec.~\ref{sec:app_annotation_pipeline}.
For each sample, the training target contains the complete five-stage reasoning sequence, followed by the three consistency scores $(S_{\mathrm{mac}}, S_{\mathrm{temp}}, S_{\mathrm{attr}})$ and the diagnostic conclusion.
We train the model with the ms-swift framework for one epoch.
The complete SFT configuration is reported in Table~\ref{tab:sft_config}.
\begin{table}[t]
\centering
\small
\setlength{\tabcolsep}{7pt}
\renewcommand{\arraystretch}{1.1}
\begin{tabular}{lc}
\toprule
\textbf{Configuration} & \textbf{Value} \\
\midrule
Base model & Qwen3-Omni-Thinking \\
Training framework & MS-SWIFT \\
Fine-tuning & Full-parameter \\
Training samples & 8k \\
Epochs & 1 \\
Global batch size & 64 \\
Peak learning rate & $5\times10^{-6}$ \\
LR scheduler & Constant with warmup \\
MoE weight & 0.001 \\
\bottomrule
\end{tabular}
\caption{Training configuration for the SFT cold-start stage.}
\label{tab:sft_config}
\end{table}

\subsection{GRPO for Human Alignment}
\label{app:grpo}

Starting from the SFT checkpoint, we further align AVC-Reward with the human-calibrated scores using GRPO.
Training is implemented with ms-swift on the 10k rollout subset constructed from the human-calibration data.

The reward combines a structure reward $r_{\mathrm{struct}}$, which enforces the required reasoning and output format, with a score-distillation reward $r_{\mathrm{distill}}$, which encourages the three predicted scores to match their human-calibrated references.

\paragraph{Structure reward.}
The structure reward $r_{\mathrm{struct}}\in[0,1]$ checks whether the model
follows the required five-stage reasoning and output format.
Its four components are summarized in Table~\ref{tab:structure_reward}.

\begin{table}[t]
\centering
\small
\setlength{\tabcolsep}{6pt}
\renewcommand{\arraystretch}{1.1}
\begin{tabular}{lr}
\toprule
\textbf{Condition} & \textbf{Reward} \\
\midrule
\texttt{<think>} content is present & +0.2 \\
Each required stage marker is present & $+0.1\times k$ ($k\leq5$) \\
All three predicted scores are present & +0.2 \\
Diagnostic conclusion is present & +0.1 \\
\bottomrule
\end{tabular}
\caption{
Components of the structure reward $r_{\mathrm{struct}}$.
}
\label{tab:structure_reward}
\end{table}

The five-stage markers are:
\begin{itemize}
\setlength{\itemsep}{0pt}
\setlength{\topsep}{2pt}
\item \texttt{step\_1\_blind\_audio\_analysis}
\item \texttt{step\_2\_deaf\_video\_analysis}
\item \texttt{step\_3\_auditory\_visual\_verification}
\item \texttt{step\_4\_visual\_auditory\_verification}
\item \texttt{step\_5\_world\_construction\_verdict}
\end{itemize}
We additionally use $r_{\mathrm{struct}}$ as a format gate:
if $r_{\mathrm{struct}}<0.4$, we set $r_{\mathrm{distill}}=0$ and truncate
the total reward to
\[
r_{\mathrm{total}}=0.5\,r_{\mathrm{struct}}.
\]
This prevents malformed outputs from receiving a high reward solely from
numerically accurate final scores.

\paragraph{Score-distillation reward.}
For structurally valid outputs, score agreement with the human-calibrated reference is measured using a Gaussian kernel:
\[
r_{\mathrm{distill}}
=
\frac{1}{3}
\sum_{i=0}^{2}
\exp\left(
-w_i(\mathrm{pred}_i-\mathrm{ref}_i)^2
\right),
\]
where
$(\mathrm{pred}_0,\mathrm{pred}_1,\mathrm{pred}_2)$
are the predicted
$(S_{\mathrm{mac}},S_{\mathrm{temp}},S_{\mathrm{attr}})$
scores and
$(\mathrm{ref}_0,\mathrm{ref}_1,\mathrm{ref}_2)$
are their calibrated human references.

We use
\[
w_0=\max(1.0,\,10-10\,\mathrm{ref}_0),
\qquad
w_1=w_2=5.0.
\]
The adaptive $w_0$ increases the penalty for rare samples with low global narrative harmony, while the temporal and attribute dimensions use fixed weights of 5.0.

For outputs that pass the structure gate, the final reward is
\[
r_{\mathrm{total}}
=
\frac{
2.0\,r_{\mathrm{struct}}
+
1.0\,r_{\mathrm{distill}}
}{3.0}.
\]
This weighting keeps the structured event-centric reasoning process as a strong training constraint while simultaneously aligning the predicted scores with human judgments.

\paragraph{GRPO configuration.}
We train with group size $G=8$ for 1,700 optimization steps.
The detailed configuration is reported in Table~\ref{tab:grpo_config}.

\begin{table}[t]
\centering
\small
\setlength{\tabcolsep}{7pt}
\renewcommand{\arraystretch}{1.1}
\begin{tabular}{lc}
\toprule
\textbf{Configuration} & \textbf{Value} \\
\midrule
Training framework & MS-SWIFT \\
Rollout samples & 10k \\
Training steps & 1,700 \\
Global batch size & 32 \\
Group size $G$ & 8 \\
Learning rate & $1\times10^{-6}$ \\
LR scheduler & Cosine decay \\
Warmup ratio & 0.05 \\
KL coefficient $\beta$ & 0.05 \\
Clipping parameter $\epsilon$ & 0.2 \\
Sampling temperature & 0.9 \\
Structure reward weight & 2.0 \\
Distillation reward weight & 1.0 \\
\bottomrule
\end{tabular}
\caption{
Training configuration for GRPO human alignment.
}
\label{tab:grpo_config}
\end{table}

\subsection{Generation-Side RL with AVC-Reward}
\label{app:generation_rl}

To evaluate AVC-Reward as an optimization signal, we build on the generation-side RL framework of OmniNFT~\citep{51omninft}.
We keep the overall training framework unchanged and replace its original audio-video reward DeSync~\citep{07desync} with AVC-Reward, using its overall score \(S_{\mathrm{AVC}}\) as the optimization signal, with reward-specific learning-rate and weight adjustments for comparable optimization strength.
This isolates the effect of the proposed reward while retaining the original generation training setup.

OmniNFT jointly optimizes audio quality, video quality, and audio-video consistency rewards; under the original weighting, AVC-Reward therefore has a weaker relative optimization effect than the original DeSync reward.
We use a two-stage schedule to compensate for this difference.
In Stage 1, we retain OmniNFT's original learning rate of $3\times10^{-5}$, set the AVC-Reward weight to 2.0, and train for 600 steps.
In Stage 2, we reduce the learning rate to $1.5\times10^{-5}$ and increase the other quality-reward weights, reducing the relative AVC-Reward weight to 1.2, and continue training for 275 steps.
This setup provides generation-side RL validation of AVC-Reward while retaining a direct controlled comparison with the original OmniNFT training framework.

\section{AVC-Benchmark Details}
\label{app:benchmark}

\subsection{Benchmark Construction and Evaluation Protocol}
\label{app:benchmark_protocol}

\paragraph{Sample selection.}
We construct AVC-Benchmark through three complementary steps.
We first perform stratified sampling based on human label configurations to balance the coverage of major consistency and failure patterns, including global narrative conflicts, temporal offsets, missing or hallucinated audio, and fine-grained attribute mismatches.
We then use ImageBind~\citep{14imagebind} to extract audio and video representations and cluster their joint embeddings, from which we sample across clusters to reduce semantic redundancy.
Finally, we add explicitly constructed negative samples to strengthen the coverage of controlled failure cases that are less frequent in naturally generated data.

The resulting benchmark contains 706 samples across different generation paradigms and data sources: 172 ground truth samples, 243 JAVG samples, 100 A2V samples, 72 V2A samples, and 119 samples from the explicit-negative construction set.
The JAVG subset contains Ovi-I2V, Ovi-T2V and Veo3~\citep{03veo3} outputs, the A2V subset is generated by an internal model, and the V2A subset is generated by HunyuanVideo-Foley.

\paragraph{Explicit negative construction.}
We construct additional controlled negatives from 50 Veo3 source clips.
Among them, 10 contain salient emotion or global narrative cues, 20 contain clear temporal relations, and 20 contain salient audio attributes.
We apply the edit operations in Table~\ref{tab:negative_construction} to produce 119 negative samples.
These edits complement naturally occurring generation failures with controlled semantic, temporal, event-existence, and fine-grained acoustic inconsistencies,  balancing the failure-case distribution and broadening coverage beyond naturally occurring generation failures.

\begin{table}[t]
\centering
\small
\setlength{\tabcolsep}{5pt}
\begin{tabular}{lr}
\toprule
\textbf{Edit operation} & \textbf{$n$} \\
\midrule
Swap audio tracks between paired clips          & 10 \\
Overlay the audio track from another clip       & 10 \\
Shift the audio by $+2$ s                       & 20 \\
Shift the audio by $-2$ s                       & 20 \\
Remove a specified audio event                  & 20 \\
Freeze the video for 1 s at an audio peak       & 19 \\
Edit sound timbre                               & 10 \\
Gradually attenuate the audio                   & 5 \\
Insert audio from another clip into static video& 5 \\
\midrule
\textbf{Total} & \textbf{119} \\
\bottomrule
\end{tabular}
\caption{
Controlled edit operations used to construct explicit negative samples.
}
\label{tab:negative_construction}
\end{table}

\paragraph{Human annotation.}
Each benchmark sample is independently evaluated by three annotators following the 39-label protocol in Appendix~\ref{sec:app_human_annotation}.
The benchmark annotators do not overlap with those used for the training set.
Rather than directly assigning scores, annotators first inspect the three dimensions independently and select all applicable labels before providing the corresponding scores.
This label-first protocol requires explicit checks of the relevant event existence, timing, attributes, and relation types, making the annotation process more traceable and reducing unconstrained score judgments.
Annotators then assign a score from 0 to 1, with a step size of 0.1, for global narrative harmony, event-level temporal synchronization, and fine-grained attribute alignment.

We use agreement among the three annotators' label records within each sample as an additional quality-control signal before constructing the final benchmark reference.
We further quantify inter-annotator agreement on the retained annotations, obtaining ICC(2,k) of 0.83, Krippendorff's $\alpha$ of 0.60, inter-annotator PLCC of 0.78, inter-annotator SRCC of 0.50, pairwise accuracy of 0.76, and $\Delta$-Pairwise@0.2 of 0.92.
The high ICC and margin-based pairwise agreement indicate stable aggregated human judgments, especially for samples with clear perceptual differences.
\paragraph{Human reference score.}

Let $s_{i,d}^{(r)}$ denote the score assigned by annotator $r$ to sample $i$
on dimension $d\in\{\mathrm{mac},\mathrm{temp},\mathrm{attr}\}$.

Since different annotators may use different scoring ranges across dimensions,
we first normalize each annotator independently. Specifically, for each annotator
$r$, we compute dimension-wise z-scores over the benchmark:

\[
z_{i,d}^{(r)}
=
\frac{s_{i,d}^{(r)}-\mu_d^{(r)}}{\sigma_d^{(r)}},
\]

where $\mu_d^{(r)}$ and $\sigma_d^{(r)}$ denote the mean and standard deviation
of annotator $r$ on dimension $d$ computed over all benchmark samples.

To further align the scales among the three dimensions for each annotator, we
apply min-max normalization over the complete three-dimensional z-score matrix
of each annotator:

\[
\tilde{z}_{i,d}^{(r)}
=
\frac{
z_{i,d}^{(r)}-\min(z^{(r)})
}{
\max(z^{(r)})-\min(z^{(r)})
},
\]

where $\min(z^{(r)})$ and $\max(z^{(r)})$ are computed over all samples and all
three dimensions for annotator $r$.

The dimension-level human reference scores are obtained by averaging the aligned
scores from three annotators:

\[
GT_{i,d}^{human}
=
\frac{1}{3}\sum_{r=1}^{3}\tilde{z}_{i,d}^{(r)} .
\]

The overall human reference score is then computed as:

\[
S_i^{human}
=
\frac{
GT_{i,\mathrm{mac}}^{human}
+
GT_{i,\mathrm{temp}}^{human}
+
GT_{i,\mathrm{attr}}^{human}
}{3}.
\]

Normalization is applied only to human reference scores to calibrate subjective
annotations across annotators. For automatic metrics, we do not introduce
additional normalization beyond each metric's original formulation. Applying
dimension-wise normalization to model outputs would additionally modify the score
distributions and aggregation behaviors of individual metrics. Since different
automatic metrics may have different output spaces, numbers of dimensions, and
aggregation strategies, applying such transformations only for a specific
evaluator would make the comparison protocol inconsistent.

For Ours, we further evaluate a z-score normalized variant as a sensitivity
analysis. As shown in Table~\ref{tab:zscore_analysis}, the difference between
raw-score and z-score aggregation is negligible, indicating that the reported
results are not sensitive to this normalization choice.

The aggregation is used only for benchmark-level overall evaluation, while the
three dimensions remain independently annotated and scored.

\begin{table}[t]
\centering
\caption{
Sensitivity analysis of score normalization for Ours on AVC-Benchmark.
}
\label{tab:zscore_analysis}
\begin{tabular}{c|ccc}
\toprule
Setting & $\Delta$-Pairwise $\uparrow$ & PLCC $\uparrow$ & SRCC $\uparrow$ \\
\midrule
Raw score aggregation & 0.728 & 0.494 & 0.244 \\
Z-score aggregation & 0.729 & 0.492 & 0.245 \\
\bottomrule
\end{tabular}
\end{table}

\paragraph{Evaluation metrics.}
We use $\Delta$-Pairwise Accuracy as the primary metric for measuring metric-to-human agreement.
For a margin $\Delta$, the valid pair set is
\[
\mathcal{P}_{\Delta} = \{(i,j): 1 \le i < j \le N,\;
|S_i^{\mathrm{human}}-S_j^{\mathrm{human}}| > \Delta\}.
\]
For an automatic metric with scores $m_i$, we compute
\[
\mathrm{Acc}_{\Delta}
= \frac{1}{|\mathcal{P}_{\Delta}|}
\sum_{(i,j)\in\mathcal{P}_{\Delta}}
\left(
\mathbb{I}\!\left[(m_i-m_j)
(S_i^{\mathrm{human}}-S_j^{\mathrm{human}})>0\right]
+ \frac{1}{2}\mathbb{I}[m_i=m_j]
\right).
\]
We use $\Delta=0.2$ as the primary setting to exclude near-tied human pairs and focus the evaluation on perceptually meaningful score differences.
Standard pairwise accuracy, SRCC, and PLCC are also computed in the complete benchmark evaluation.

\paragraph{PLCC with five-parameter logistic mapping.}
Following common practice in subjective quality assessment~\citep{sheikh2003image},
we apply a nonlinear logistic regression before computing PLCC.
Human perceptual judgments are often not linearly related to objective prediction
scores due to perceptual saturation and scale differences.
Therefore, a mapping is first learned to align the prediction
scale with the human score space.

Specifically, for each predictor and evaluation dimension, we fit a five-parameter
logistic function on the same evaluation subset:

\begin{equation}
f(x)=b_1
\left(
0.5-\frac{1}{1+\exp(b_2(x-b_3))}
\right)
+b_4x+b_5 ,
\end{equation}

where $x$ denotes the predicted score and $f(x)$ is the calibrated prediction.
The parameters $\{b_1,b_2,b_3,b_4,b_5\}$ are optimized by nonlinear least squares
fitting between predictions and human ground-truth scores.

The reported PLCC is then computed as the Pearson correlation coefficient between
the calibrated prediction $f(x)$ and the human score $y$:

\begin{equation}
\mathrm{PLCC}
=
\frac{
\mathrm{Cov}(f(x),y)
}{
\sigma_{f(x)}\sigma_y
}.
\end{equation}

If the nonlinear fitting fails, we fall back to Pearson correlation computed on
the original prediction scores. Note that SRCC and $\Delta$-Pairwise are computed
using the original predictions without logistic calibration, since they only
depend on score ordering.

\subsection{Benchmark Statistics}
\label{app:benchmark_statistics}

\begin{figure}[t]
    \centering
    \includegraphics[width=\linewidth]{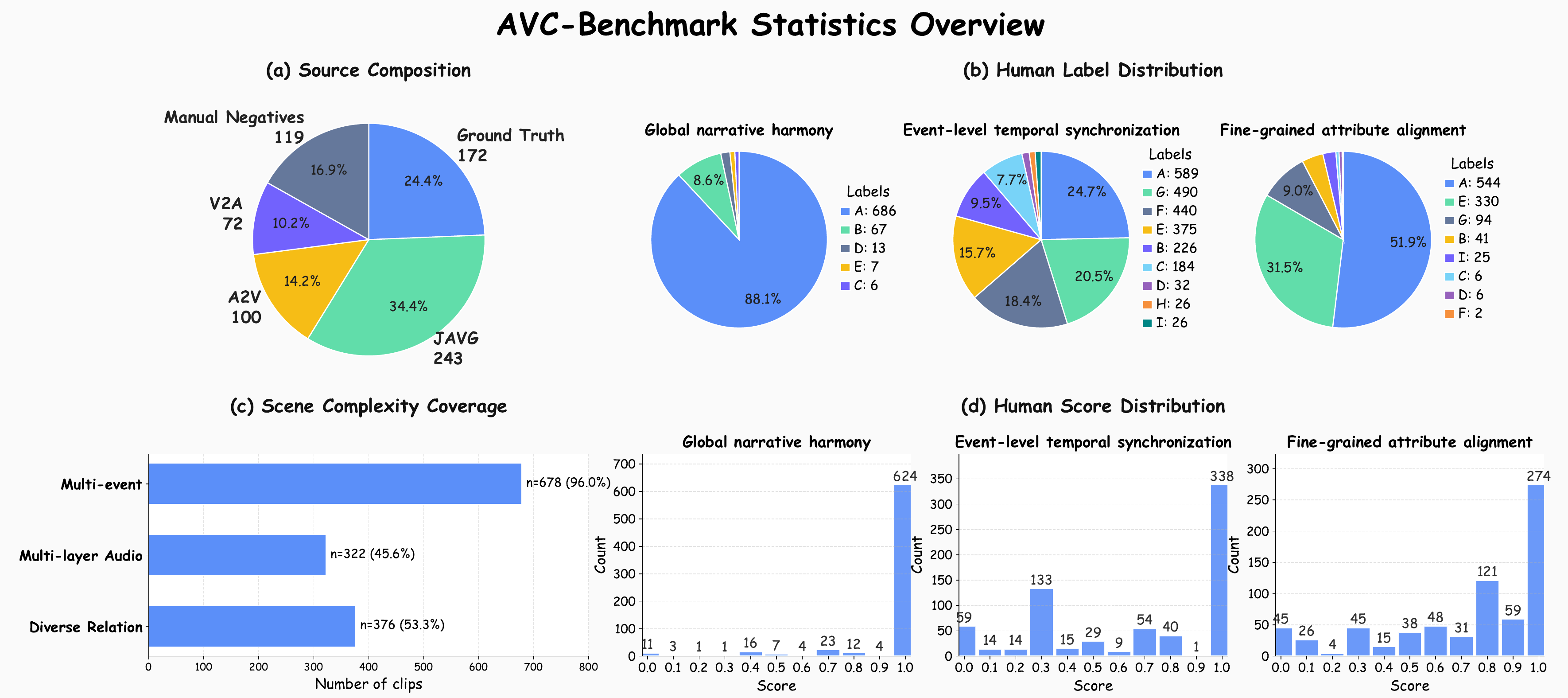}
    \caption{
    Statistics of AVC-Benchmark, including source composition, human-label    distribution, scene-complexity coverage, and human-score distribution.
    The three complexity subsets are non-exclusive.
    }
    \label{fig:benchmark_statistics}
\end{figure}

Figure~\ref{fig:benchmark_statistics} summarizes the composition of AVC-Benchmark.
The 706 samples cover ground truth, JAVG, A2V, V2A, and explicitly constructed negative data.
Based on the human annotations, 678 samples (96.0\%) exhibit multi-event composition, 322 (45.6\%) contain multi-layer audio, and 376 (53.3\%) contain diverse cross-modal relations.
The three complexity subsets are non-exclusive, allowing a sample to exhibit multiple forms of complexity simultaneously.
Together with the 39-label and human-score distributions, these statistics show that AVC-Benchmark covers different generation settings, failure patterns, and relation complexities.

\paragraph{Pair counts under different margins.}

\begin{table*}[t]
\centering
\small
\setlength{\tabcolsep}{6pt}
\begin{tabular}{lcccccccc}
\toprule
$\Delta$
& 0.00 & 0.05 & 0.10 & 0.15 & \textbf{0.20} & 0.25 & 0.30 & 0.50 \\
\midrule
Valid pairs
& 232,716 & 166,029 & 116,199 & 86,224
& \textbf{70,352} & 60,485 & 53,579 & 31,512 \\
\bottomrule
\end{tabular}
\caption{
Number of valid sample pairs at different $\Delta$-Pairwise margins using the
z-score-normalized human reference.
}
\label{tab:delta_pair_counts}
\end{table*}
Table~\ref{tab:delta_pair_counts} reports the number of valid sample pairs under different margins after constructing the z-score-normalized human reference.
At the primary threshold $\Delta=0.2$, 70,352 valid pairs remain for evaluation.

\subsection{Fine-Grained Results by Scene Complexity}
\label{app:benchmark_finegrained}

\begin{table*}[t]
\centering
\small
\setlength{\tabcolsep}{7pt}
\begin{tabular}{lccc}
\toprule
\textbf{Method}
& \textbf{Multi-event}
& \textbf{Multi-layer Audio}
& \textbf{Diverse Relation} \\
& \textbf{(678)}
& \textbf{(322)}
& \textbf{(376)} \\
\midrule
AV-Align                    & 0.654 & 0.681 & 0.670 \\
DeSync                      & 0.587 & 0.582 & 0.567 \\
ImageBind                   & 0.626 & 0.658 & 0.632 \\
JavisScore                  & 0.615 & 0.659 & 0.636 \\
AVHScore                    & 0.597 & 0.635 & 0.611 \\
PEAVS                       & 0.504 & 0.590 & 0.564 \\
AVSyncEvaluator             & 0.551 & 0.457 & 0.455 \\
AV-Consistency Evaluator    & 0.659 & 0.688 & 0.653 \\
AGAV-Rater                  & \underline{0.661} & 0.629 & 0.641 \\
Gemini-3.1-Pro              & 0.612 & 0.622 & 0.608 \\
Qwen3-Omni-Instruct         & 0.591 & 0.620 & 0.593 \\
Qwen3-Omni-Thinking         & 0.597 & \underline{0.696} & \underline{0.698} \\
\midrule
AVC-Reward                  & \textbf{0.715} & \textbf{0.747} & \textbf{0.703} \\
\midrule
Human Upper Bound           & 0.923 & 0.944 & 0.934 \\
\bottomrule
\end{tabular}
\caption{
$\Delta$-Pairwise@0.2 on the three non-exclusive scene-complexity subsets.
The best and second-best automatic metrics are shown in bold and underlined, respectively. Human Upper Bound is excluded from ranking.
}
\label{tab:complexity_results}
\end{table*}
We further evaluate metric-to-human agreement on the three scene-complexity subsets derived from the human annotations.
The subsets are non-exclusive and correspond to the three properties motivating our evaluation design: multi-event composition, multi-layer audio, and diverse cross-modal relations.

Table~\ref{tab:complexity_results} shows that AVC-Reward achieves the highest $\Delta$-Pairwise@0.2 on all three complexity subsets.
The margins are clearer on multi-event and multi-layer-audio samples, where AVC-Reward improves from the second-best results of 0.661 and 0.696 to 0.715 and 0.747, respectively.
On diverse-relation samples, AVC-Reward also gives the best pairwise agreement, although the margin over Qwen3-Omni-Thinking is smaller (0.703 vs.\ 0.698).
These results show that the event-centric evaluation is particularly effective for samples involving multiple events and layered audio, while retaining its advantage when audio and video are related through off-screen, narrative, or other non-trivial relations.

\subsection{Dimension Correspondence Analysis}

\begin{table}[t]
\centering
\caption{
Cross-dimensional permutation analysis for validating dimension correspondence.
Values denote AUC decrease after permuting each predicted dimension.
}
\label{tab:dimension_correspondence}
\begin{tabular}{c|ccc}
\toprule
Human target &
$\Delta$ Macro &
$\Delta$ Temporal &
$\Delta$ Fine-grained
\\
\midrule
Macro
& \textbf{0.367}
& 0.026
& -0.032
\\

Temporal
& -0.009
& \textbf{0.078}
& 0.038
\\

Fine-grained
& -0.004
& 0.039
& \textbf{0.102}
\\

\bottomrule
\end{tabular}
\end{table}
To verify whether the three predicted scores correspond to their intended
consistency dimensions, we perform a dimension correspondence analysis based on
human-annotated issue labels.

Instead of directly measuring the correlation between predicted dimension scores
and human dimension scores, we formulate this analysis as a dimension-specific
issue classification task. Human dimension scores are mainly designed for
overall perceptual evaluation, and the three dimensions may have different
levels of granularity. Therefore, compared with continuous scores, human issue
labels provide a more direct signal for evaluating whether each predicted
dimension corresponds to its intended consistency dimension.

Specifically, we construct dimension-specific binary targets from human-annotated issue labels in the benchmark, where samples are separated into
positive and negative cases according to whether they contain the corresponding
types of consistency issues. We then use the three predicted scores from Ours as
input features and evaluate a classifier with cross-validated AUC.

To measure the contribution of each predicted dimension, we randomly permute one
prediction dimension while keeping the remaining dimensions unchanged. The
performance decrease is calculated as:

\[
\Delta AUC =
AUC_{\mathrm{original}}
-
AUC_{\mathrm{permuted}} .
\]

A larger $\Delta AUC$ indicates that the corresponding predicted dimension
provides more information for identifying the target consistency issue category.
Results are averaged over repeated cross-validation runs. As shown in
Table~\ref{tab:dimension_correspondence}, the largest performance drop is
consistently observed when permuting the corresponding dimension, supporting the
correspondence between model outputs and human-defined consistency dimensions.

\section{More Case Studies}
\label{app:case}
\begin{figure*}[t]
    \centering
    \includegraphics[width=\textwidth]{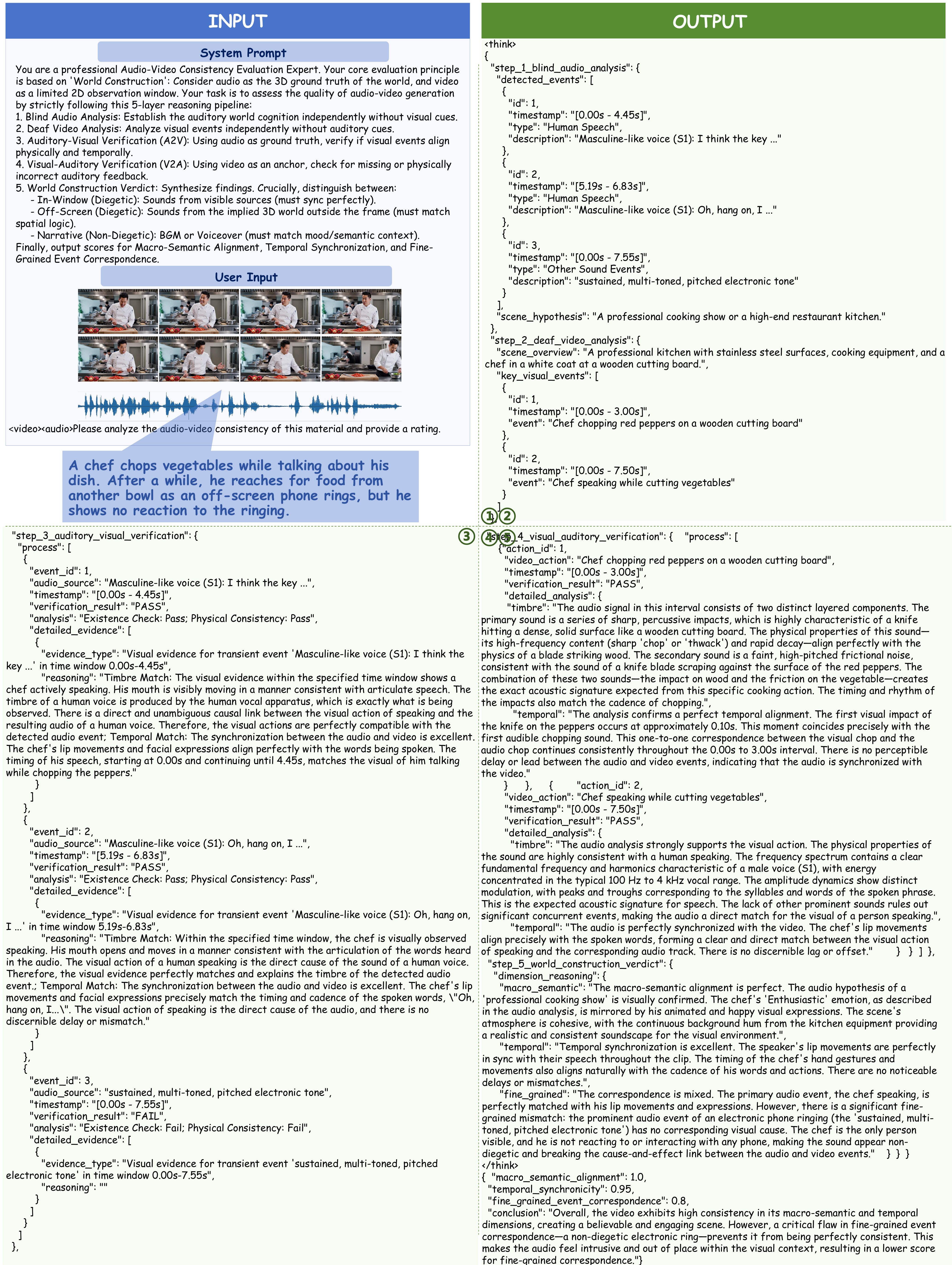}
    \caption{Full input--output example of AVC-Reward. For visualization purposes, the semantically empty lines in Stage 4 and Stage 5 are compressed due to space limitations.}
    \label{fig:appendix_case_prompt}
\end{figure*}
\begin{figure*}[t]
    \centering
    \includegraphics[width=\textwidth]{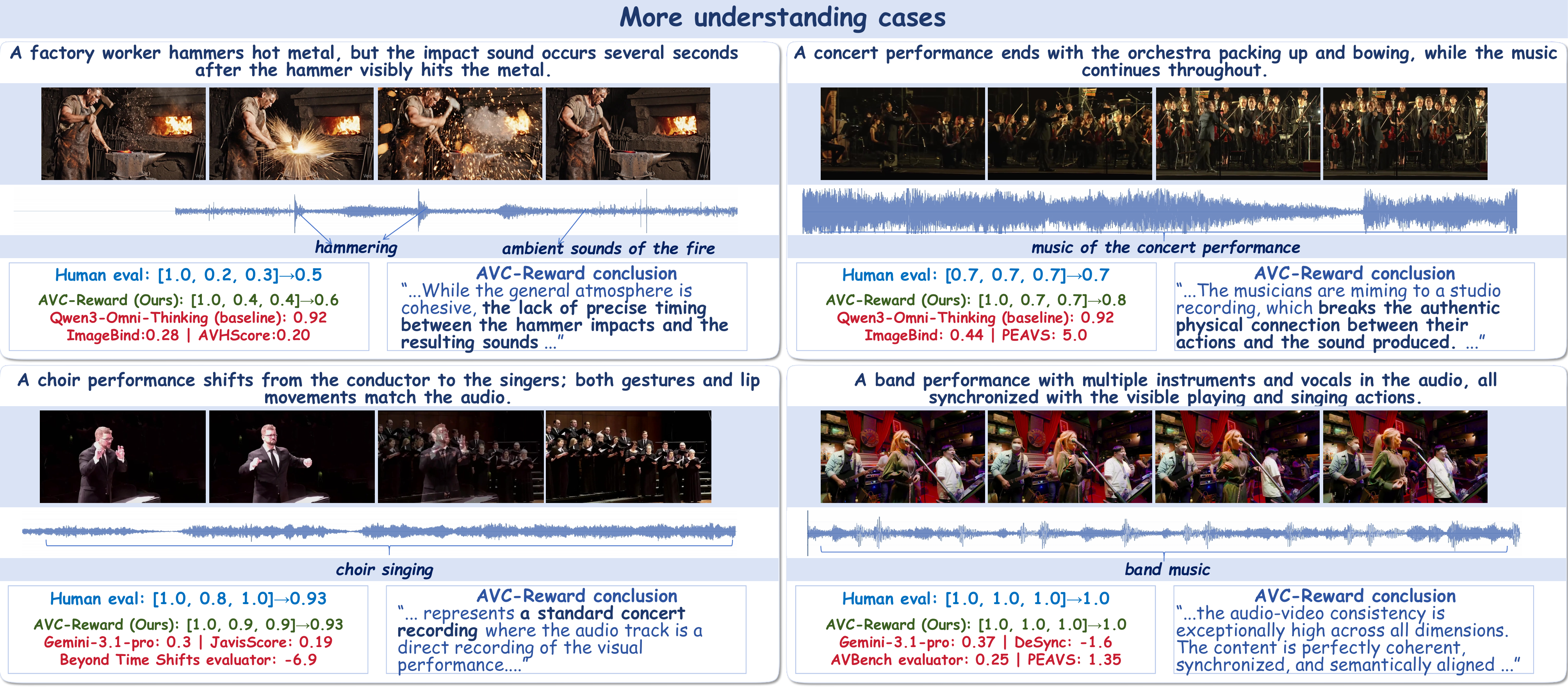}
    \caption{Additional evaluation cases of AVC-Reward.}
    \label{fig:appendix_case_3_2}
\end{figure*}
\begin{figure*}[t]
    \centering
    \includegraphics[width=\textwidth]{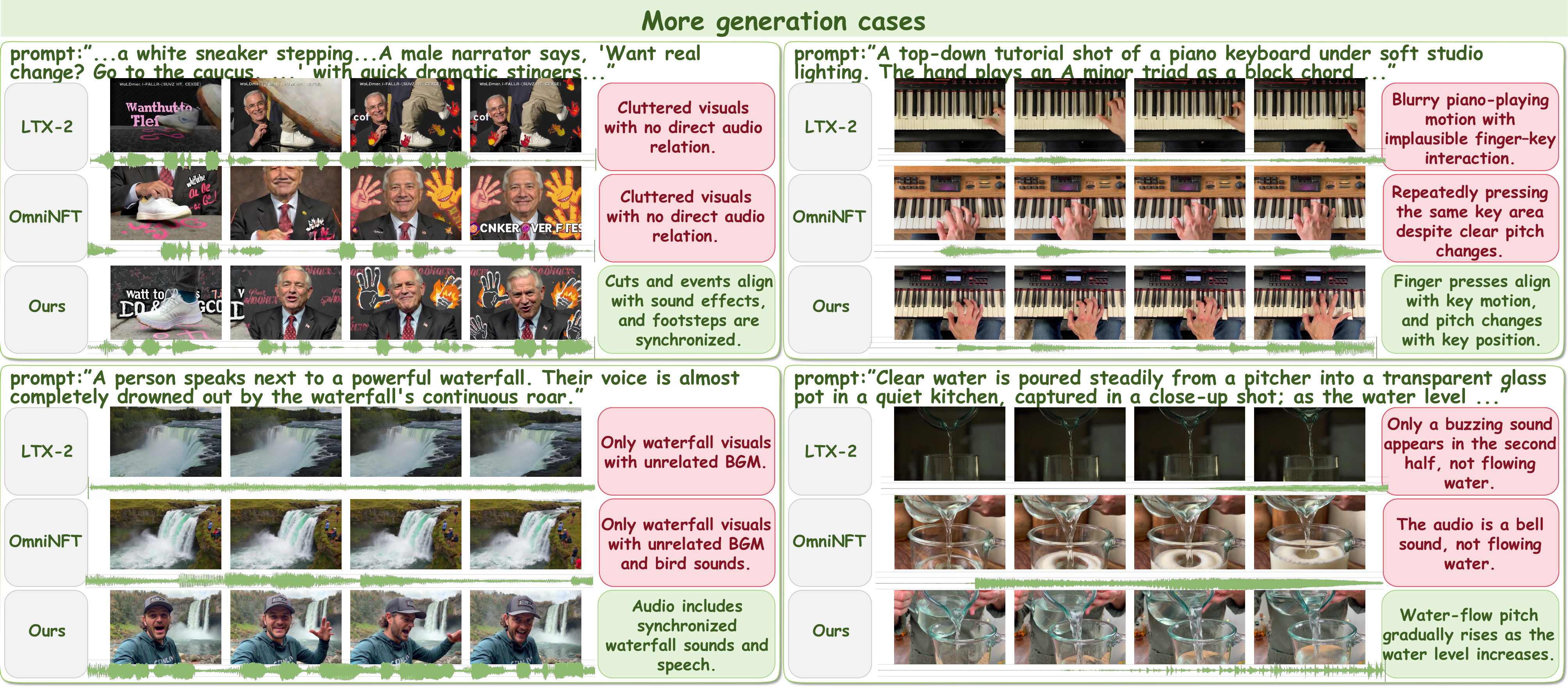}
    \caption{Additional generation cases after AVC-Reward optimization.}
    \label{fig:appendix_case_3_3}
\end{figure*}
\paragraph{Full AVC-Reward input--output.}
Fig.~\ref{fig:appendix_case_prompt} shows a complete input--output example of AVC-Reward.
The clip contains a coherent kitchen scene with several events, including chef speech, chopping, and an off-screen phone ring.
The phone ring can plausibly originate outside the frame, but the visible chef shows no reaction or other visual evidence supporting its interaction with the scene.
AVC-Reward first analyzes the audio and video independently to extract their scenes and major events, and then performs audio-to-video and video-to-audio verification for the extracted events.
The final stage combines the scene-level information and cross-modal evidence to judge the three consistency dimensions.
The complete reasoning process is contained in the \texttt{<think>} block, followed by the three scores and a diagnostic conclusion.

\paragraph{Additional evaluation cases.}
Fig.~\ref{fig:appendix_case_3_2}, together with the evaluation cases in Fig.~\ref{fig:cases}, further shows how AVC-Reward handles different forms of complexity in generated audio-video content.
The two-person conversation case in Fig.~\ref{fig:cases} tests multi-event disentanglement: two people are visible, but the audio provides clear speech evidence for only one of them.
The puppet case tests multi-layer audio, where character speech, ambient factory sounds, and background music play different roles in the scene.

The additional cases extend this evaluation to several other situations.
The band case contains multiple instruments and vocals at the same time, testing concurrent event verification.
The choir case changes from a conductor to the singers and requires the evaluator to verify different valid audio-video relations across the clip.
The hammer case tests precise event-level temporal verification by introducing a clear delay between the visible impact and its sound.
The concert-ending case tests fine-grained visual evidence for sound production: the performance has ended and the orchestra begins packing up and bowing, although the performance audio continues.
Across these cases, AVC-Reward remains close to human judgments and provides conclusions that identify the relevant event or relation.
The positive choir and band examples further show that complex but valid relations can still receive high consistency judgments.

\paragraph{Additional generation cases.}
Fig.~\ref{fig:appendix_case_3_3}, together with the generation cases in Fig.~\ref{fig:cases}, shows how the same event-level evaluation capabilities are reflected in generation after using AVC-Reward for optimization.
In the dog--cat case in Fig.~\ref{fig:cases}, the optimized model better preserves sequential multi-event composition, with the dog and cat opening their mouths for their respective sounds.
In the cooking case, it better handles multi-layer audio by retaining narration while also generating visible chopping with synchronized chopping sounds.

The additional cases cover further generation complexities.
In the waterfall case, the optimized model preserves concurrent waterfall sound and human speech.
In the election-ad case, visual cuts, sound effects, footsteps, and narration form different event-dependent relations that are more consistently generated.
The water-pouring case shows finer audio attributes, where the water-flow sound changes with the evolving visual event.
The piano case shows finer visual evidence for sound production, with finger presses, key motion, and pitch changes forming a more coherent relation.
Together, these examples qualitatively mirror the capabilities examined by AVC-Reward: finer checks of individual events, their cross-modal relations, audio attributes, and visible sound-producing evidence lead to more complete event generation and more consistent audio-video relations.

Together, these cases cover the main capabilities examined on both the evaluation and generation sides, including multi-event composition, concurrent events, complex cross-modal relations, multi-layer audio, precise temporal verification, and fine-grained audio or visual evidence.

\paragraph{Terminology note.} The original prompts and outputs above retain the terminology used during data construction. Macro-Semantic Alignment, Temporal Synchronicity, and Fine-Grained Event Correspondence correspond to Global Narrative Harmony, Event-Level Temporal Synchronization, and Fine-Grained Attribute Alignment in the main text, respectively.

\section{Human Evaluation Protocol and Results}
\label{app:human_eval}

\begin{table*}[t]
\centering
\caption{Normalized human support rates and pairwise win rates.  Pairwise win rates exclude double selections containing both compared systems and Tie responses.}
\label{tab:human_eval_results}
\tablefontsize
\resizebox{\textwidth}{!}{\begin{tabular}{llccccc}
\toprule
Benchmark & Question & Ours & OmniNFT & LTX-2 & Ours$>$OmniNFT & Ours$>$LTX-2 \\
\midrule
AVGenBench & Consistency & 47.8 & 24.4 & 27.8 & 69.1 & 65.5 \\
AVGenBench & Overall & 46.2 & 27.5 & 26.4 & 66.0 & 64.5 \\
AV-Phys-Bench & Consistency & 50.0 & 27.8 & 22.2 & 73.1 & 71.4 \\
AV-Phys-Bench & Overall & 49.5 & 26.2 & 24.3 & 72.2 & 68.6 \\
VABench & Consistency & 46.4 & 34.1 & 19.6 & 63.5 & 72.3 \\
VABench & Overall & 44.8 & 36.0 & 19.2 & 57.7 & 70.0 \\
\bottomrule
\end{tabular}
}\end{table*}

\begin{table*}[t]
\centering
\caption{Normalized support rates and pairwise win rates on the fully consistent subset.}
\label{tab:human_eval_agreement}
\tablefontsize
\resizebox{\textwidth}{!}{\begin{tabular}{lcccc}
\toprule
Question & Ours & OmniNFT & LTX-2 & Ours$>$OmniNFT / Ours$>$LTX-2 \\
\midrule
Audio-video consistency & 64.3 & 7.1 & 28.6 & 90.0 / 72.7 \\
Overall quality & 56.3 & 18.8 & 25.0 & 75.0 / 69.2 \\
\bottomrule
\end{tabular}
}\end{table*}

\begin{figure*}[t]
\centering
\includegraphics[width=0.96\textwidth]{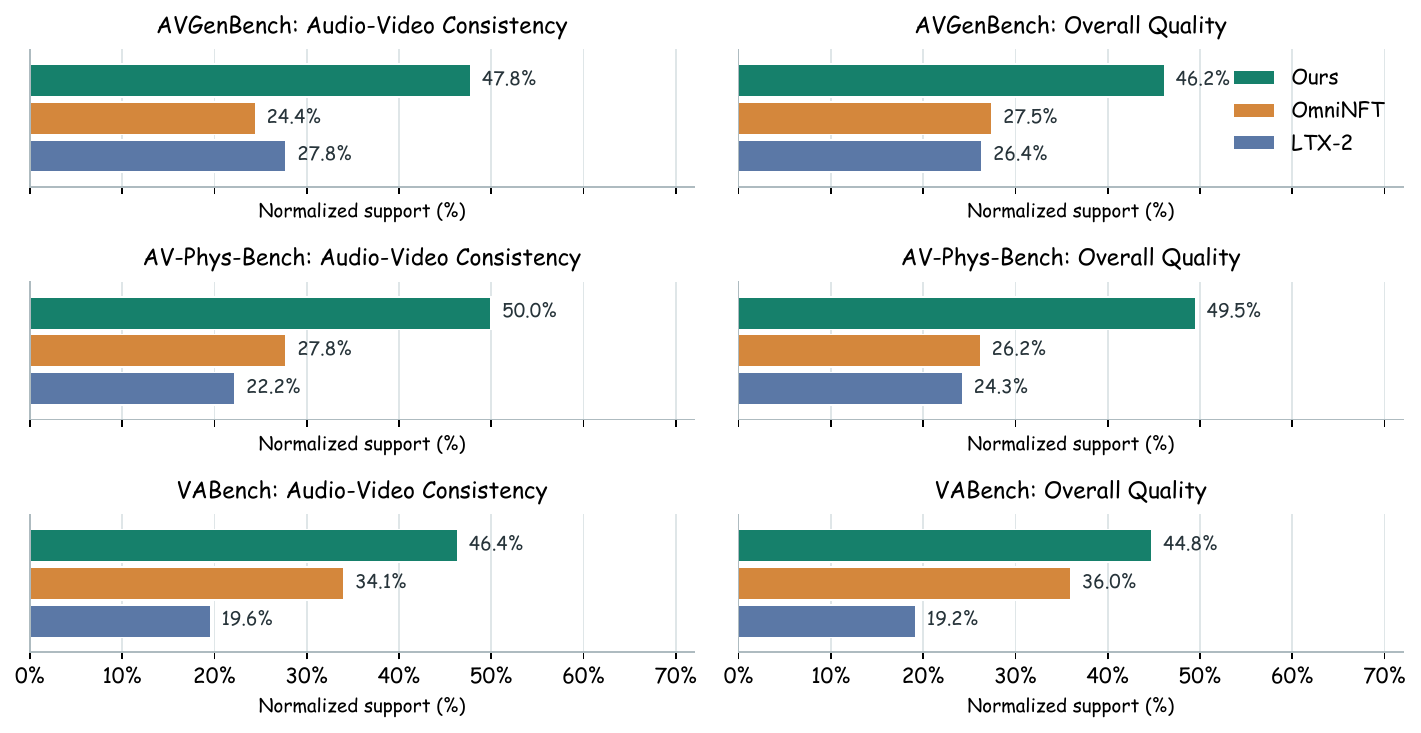}
\caption{Benchmark-wise normalized human support rates. Each row corresponds to one benchmark, and the two columns report audio-video consistency and overall quality.}
\label{fig:human_eval_appendix_benchmarks}
\end{figure*}
\subsection{Sampling and Annotation Protocol}
\label{app:human_eval_protocol}

We randomly sample 100 prompts from the three generation benchmarks with fixed coverage: 30 prompts from AVGenBench, 30 from AV-Phys-Bench, and 40 from VABench. For each prompt, we generate or collect one sample from each of LTX-2, OmniNFT, and Ours. Each case therefore contains three videos with the same English prompt. The three videos are shown as anonymous options A, B, and C in a randomized order, and model identities are hidden from annotators.

Ten annotators with at least a bachelor's degree independently evaluate 30 cases each. The assignment is balanced so that every case is evaluated by exactly three annotators. Annotators answer two questions:

\begin{enumerate}
    \item \textbf{Audio-video consistency.} Which sample better matches sound events to the visible action in timing and attributes? The judgment also considers whether narration, off-screen sound, ambient sound, and background music are appropriate for the scene.
    \item \textbf{Overall quality.} Which sample is better when audio quality, visual quality, and audio-video consistency are considered together?
\end{enumerate}

The default response is one selected sample. An annotator may select two samples only when they are both clearly better than the third and are difficult to distinguish from each other. \texttt{Tie} is selected when the three samples are approximately equivalent. A selection of two samples is retained as a two-element set; \texttt{Tie} is retained as a separate category.

\subsection{Aggregation Rules}
\label{app:human_eval_aggregation}

For a model $m$, let $S_m$ be the number of responses in which $m$ is selected. A two-element response contributes one support to each selected model, while a \texttt{Tie} contributes no support. We report the normalized support rate
\begin{equation}
    R_m = \frac{S_m}{\sum_{m'\in\mathcal{M}} S_{m'}},
    \qquad
    \mathcal{M}=\{\mathrm{Ours},\mathrm{OmniNFT},\mathrm{LTX\text{-}2}\}.
\end{equation}

For the pairwise comparison between Ours and a baseline $b$, a response counts as an Ours win when Ours is selected and $b$ is not selected. It counts as a baseline win when $b$ is selected and Ours is not selected. Responses selecting both systems or \texttt{Tie} are excluded from that pairwise denominator. The pairwise win rate is therefore
\begin{equation}
    P_{\mathrm{Ours}>b}
    = \frac{W_{\mathrm{Ours}>b}}
    {W_{\mathrm{Ours}>b}+W_{b>\mathrm{Ours}}}.
\end{equation}

The mean pairwise Jaccard agreement is 42.3\% for audio-video consistency and 44.3\% for overall quality. This medium overlap is acceptable for a comparison task with substantial subjectivity and allowed near-ties. Because the task is subjective, we additionally construct a fully consistent subset in which all three annotators select the same set for the same prompt and question, and report its model support rates below.

\subsection{Benchmark-wise Results}
\label{app:human_eval_results}
Table~\ref{tab:human_eval_results} reports support rates and the pairwise win rates of Ours against each baseline.  The results use all 300 completed judgments.
As shown in Fig.~\ref{fig:human_eval_appendix_benchmarks}, Ours has an advantage on both questions in every benchmark, with the clearest advantage on AV-Phys-Bench.

\subsection{Fully Consistent Subset Results}
\label{app:human_eval_agreement}

As shown in Table~\ref{tab:human_eval_agreement}, on the fully consistent subset, Ours obtains 64.3\% support for audio-video consistency versus 7.1\% for OmniNFT and 28.6\% for LTX-2, with pairwise win rates of 90.0\% and 72.7\%.  For overall quality, Ours obtains 56.3\% support versus 18.8\% and 25.0\%, with pairwise win rates of 75.0\% and 69.2\%. These results show that the preference advantage of Ours becomes clearer on the high-agreement subset.

\end{document}